\pdfoutput=1

\documentclass[11pt]{article}

\usepackage[]{acl}

\usepackage{times}
\usepackage{latexsym}

\usepackage[T1]{fontenc}

\usepackage[utf8]{inputenc}

\usepackage{microtype}

\usepackage{inconsolata}

\usepackage{times}
\usepackage{latexsym}
\usepackage{amsfonts}

\usepackage[T1]{fontenc}

\usepackage[utf8]{inputenc}

\usepackage{microtype}

\usepackage{inconsolata}
\usepackage{graphicx}
\usepackage{booktabs}
\usepackage{caption}
\usepackage{times}
\usepackage{latexsym}
\usepackage{svg}
\usepackage{amsmath}
\usepackage{tcolorbox}\tcbuselibrary{skins}
\usepackage{subcaption}
\usepackage{multirow}
\usepackage{array}
\usepackage{makecell}
\usepackage{float}
\usepackage{booktabs}
\usepackage{times}
\usepackage{latexsym}
\usepackage{graphicx}
\usepackage{url}
\usepackage{xcolor,colortbl}
\usepackage{soul}
\usepackage{color}
\usepackage{CJKutf8}
\usepackage{enumitem}
\usepackage{cleveref}

\crefformat{section}{\S#2#1#3}
\crefformat{subsection}{\S#2#1#3}
\crefformat{subsubsection}{\S#2#1#3}

\newcommand{\hlc}[2][yellow]{{%
    \colorlet{foo}{#1}%
    \sethlcolor{foo}\hl{#2}}%
}

\usepackage{hyperref}
\usepackage{url}
\usepackage{soul}
\usepackage{makecell}
\usepackage{threeparttable,subcaption}
\usepackage{tikz}
\usepackage{multirow}
\usepackage{booktabs,wrapfig}
\usepackage{amsthm}
\usepackage{setspace}
\usepackage{tcolorbox}
\newcommand{\fon}[1]{\fontfamily{#1}\selectfont} 

\def\eg{e.g.,}

\def\taska{Step A}
\def\taskb{Step B}
\def\taskc{Step C}

\def\chatgpt{\texttt{gpt-3.5-turbo}}
\def\gptfour{\texttt{gpt-4}}
\def\llama{\texttt{llama-2}}
\def\methodname{\texttt{Bridge}}
\def\numAnnotations{700}
\def\numTrainAnnotations{420}
\def\numValidationAnnotations{70}
\def\numTestAnnotations{210}

\def\guessLabel{\texttt{guess}}
\def\guessDescription{The student does not seem to understand or guessed the answer.}

\def\misinterpretLabel{\texttt{misinterpret}}
\def\misinterpretDescription{The student misinterpreted the question.}

\def\carelessLabel{\texttt{careless}}
\def\carelessDescription{The student made a careless mistake.}

\def\rightIdeaLabel{\texttt{right-idea}}
\def\rightIdeaDescription{The student has the right idea, but is not quite there.}

\def\impreciseLabel{\texttt{imprecise}}
\def\impreciseDescription{The student's answer is not precise enough or the tutor is being too picky about the form of the student's answer.}

\def\notSureLabel{\texttt{not-sure}}
\def\notSureDescription{Not sure, but I'm going to try to diagnose the student.}

\def\naLabel{\texttt{N/A}}
\def\naDescription{None of the above, I have a different description.}

\newcommand\graycell{\cellcolor[rgb]{0.9,0.9,0.9}}
\newcommand\yellowcell{\cellcolor[rgb]{1,0.843,0}}

\usepackage{array}
\newcolumntype{P}[1]{>{\centering\arraybackslash}p{#1}}

\tcbset{taska/.style={
    enhanced,
    size=fbox,
    boxrule=3pt,
    arc=2mm,
    auto outer arc,
    left=10pt,
    right=10pt,
    top=10pt,
    bottom=10pt,
    fontupper=\fon{cmtt}, 
    colback=CB_pear!10,
    colframe=CB_pear!25,
    coltitle=CB_pear!20!black, 
}}

\tcbset{taskb/.style={
    enhanced,
    size=fbox,
    boxrule=3pt,
    arc=2mm,
    auto outer arc,
    left=10pt,
    right=10pt,
    top=10pt,
    bottom=10pt,
    fontupper=\fon{cmtt}, 
    colback=CB_lightCyan!15,
    colframe=CB_lightCyan!30,
    coltitle=CB_lightCyan!25!black, 
}}

\tcbset{taskc/.style={
    enhanced,
    size=fbox,
    boxrule=3pt,
    arc=2mm,
    auto outer arc,
    left=10pt,
    right=10pt,
    top=10pt,
    bottom=10pt,
    colback=CB_pink!15,
    colframe=CB_pink!30,
    coltitle=CB_pink!25!black, 
    fontupper=\fon{cmtt}, 
}}

\usepackage{tikz}

\definecolor{CB_pear}{HTML}{BBCC33}
\definecolor{CB_pink}{HTML}{FFAABB}
\definecolor{CB_lightCyan}{HTML}{99DDFF}
\definecolor{CB_gray}{HTML}{DDDDDD}
\definecolor{CB_orange}{HTML}{EE8866}

\newcommand{\emojistudent}{
    \includegraphics[scale=0.03]{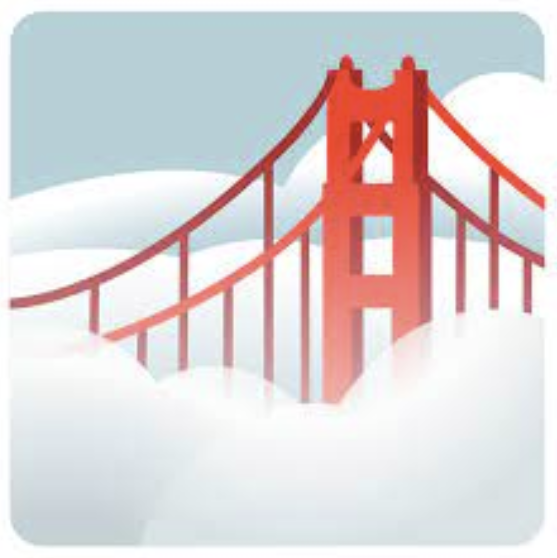}}

\title{\hspace{-0.6em}\emojistudent{} 
Bridging the Novice-Expert Gap via Models of Decision-Making: \\
A Case Study on Remediating Math Mistakes
}

\author{\vspace{.5em}{\bf Rose E. Wang}\quad {\bf Qingyang Zhang}\quad {\bf Carly Robinson}\quad \\ \vspace{.5em} {\bf Susanna Loeb}\quad {\bf Dorottya Demszky}\\ \vspace{.5em} Stanford University\\ \texttt{rewang@cs.stanford.edu, ddemszky@stanford.edu}}

\begin{document}
\maketitle
\begin{abstract}
Scaling high-quality tutoring remains a major challenge in education.
Due to growing demand, many platforms employ novice tutors who, unlike experienced educators, struggle to address student mistakes and thus fail to seize prime learning opportunities.
Our work explores the potential of large language models (LLMs) to close the novice-expert knowledge gap in remediating math mistakes.
We contribute \methodname{}, a method that uses cognitive task analysis to translate an expert's latent thought process into a decision-making model for remediation.
This involves an expert identifying (A) the student's error, (B) a remediation strategy, and (C) their intention before generating a response.
We construct a dataset of 700 real tutoring conversations, annotated by experts with their decisions.
We evaluate state-of-the-art LLMs on our dataset and find that the expert's decision-making model is critical for LLMs to close the gap: 
responses from GPT4 with expert decisions (\eg{} ``simplify the problem'') are +76\% more preferred than without.
Additionally, context-sensitive decisions are critical to closing pedagogical gaps: 
random decisions decrease GPT4's response quality by -97\% than expert decisions.
Our work shows the potential of embedding expert thought processes in LLM generations to enhance their capability to bridge novice-expert knowledge gaps. Our dataset and code can be found at: \url{https://github.com/rosewang2008/bridge}.
\end{abstract}

\section{Introduction}

\begin{figure*}[t]
    \centering
    \includegraphics[width=\linewidth]{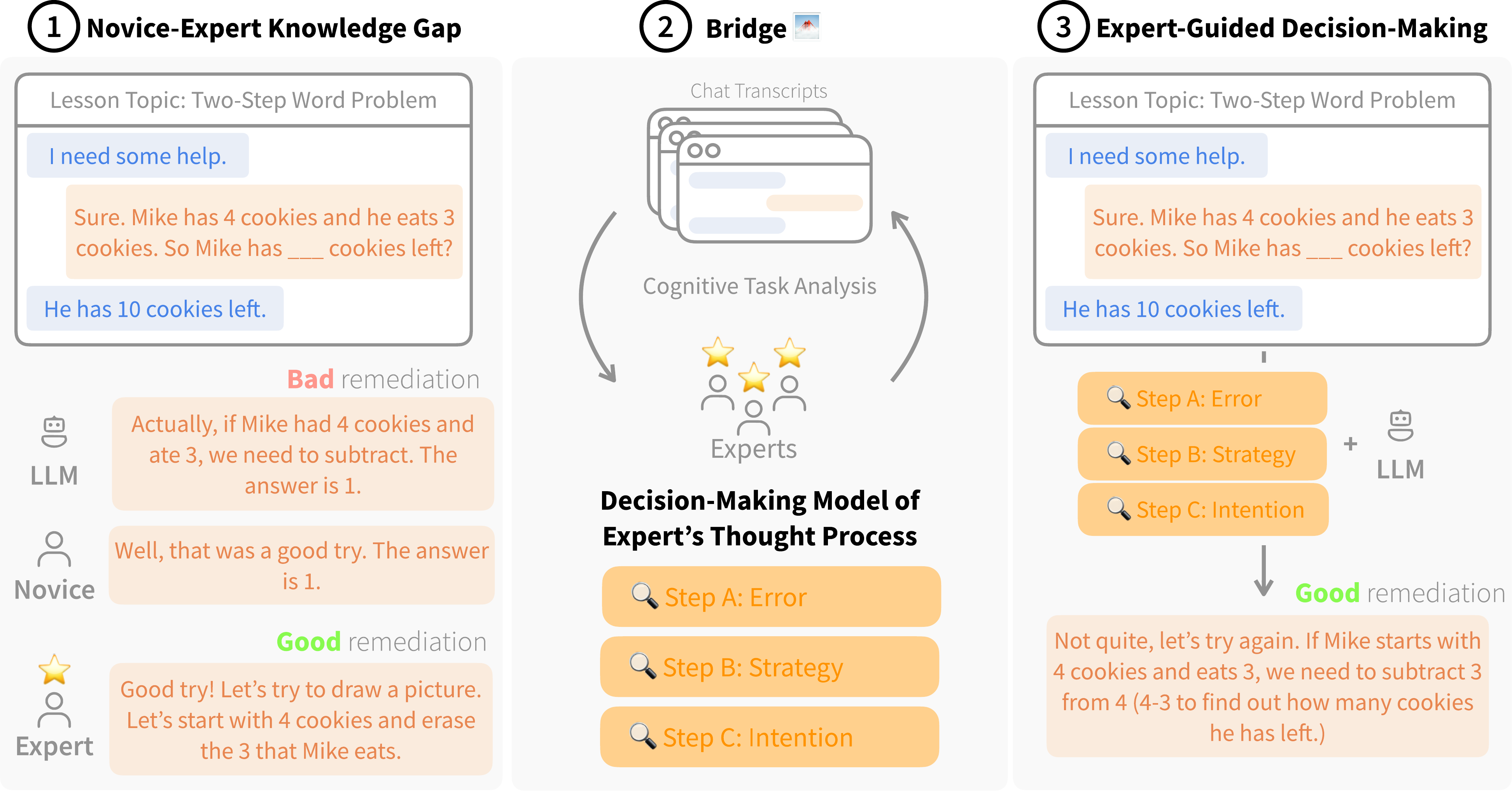}
    \caption{
    \textbf{\raisebox{.5pt}{\textcircled{\raisebox{-.8pt} {1}}} Closing the knowledge gap at scale.} 
    LLMs and novice tutors lack the pedagogical knowledge to engage with student mistakes, yet they are readily available for 1:1 tutoring. 
    Experts like experienced teachers have the pedagogical knowledge, but are hard to scale.
    \textbf{\raisebox{.5pt}{\textcircled{\raisebox{-.8pt} {2}}} How do we model the expert's thought process?} 
    Our work builds \methodname{} which leverage cognitive task analysis to translate the latent thought process of experts into a decision-making model. 
    \textbf{\raisebox{.5pt}{\textcircled{\raisebox{-.8pt} {3}}} Applying \methodname{} with LLMs.} To bridge the knowledge gap, we scale the expert's knowledge with LLMs using the expert-guided decision-making model.
    }
    \label{fig:method}
\end{figure*}  

Human tutoring plays a critical role in accelerating student learning, and is one of the primary ways to combat pandemic-related learning losses 
\citep{fryer2020high, nickow2020impressive,robinson2021high,department2021,nssa2022}. 
To accommodate the growing demand for tutoring, many tutoring providers engage novice tutors. 
While novice tutors may exercise the domain knowledge, they often lack the specialized training of professional educators in interacting with students. 
However, research suggests that novices with proper training can be effective tutors \citep{nickow2020impressive}. 

Responding to student mistakes in real-time is a critical area where novice tutors tend to struggle. 
Mistakes are prime learning opportunities to address misconceptions \citep{boaler2013ability}, but effective responses involve pedagogical expertise in engaging with student's thinking and building positive rapport \citep{roorda2011influence, pianta2016teacher, shaughnessy2021think, robinson2022framework}.
Novices typically learn from experts to understand the expert's thought process however hiring experienced educators to provide timely feedback is resource-intensive \citep{KraftBlazarHogan2018,kelly2020using}. 

One potential solution is the use of automated tutors \citep{graesser2004autotutor}. 
With recent advances in large language models (LLMs), this approach has gained even more interest \citep{khan_academy_2023}. 
However their ability to remediate is yet to be evaluated.  
Prior work suggests several shortcomings with LLMs, including lacking reliable subject and pedagogical knowledge \citep{frieder2023mathematical, wang2023a,Singer_2023}, that can be mitigated using explicitly thought processes such as through chain-of-thought prompting \citep{wei2022chain}.

To address these challenges, our work makes several key contributions.
First, we build \textbf{\methodname{}, a method that leverages cognitive task analysis to elicit the latent thought processes of experts}. We apply \texttt{Bridge} to remediation where we collaborate extensively with experienced math educators to translate their thought process into a decision-making model.
\methodname{} breaks down the experts' thought process: illustrated in Figure~\ref{fig:method}, 
\taska{} is to infer the student's error (\eg{} the student guessed); 
\taskb{} is to determine the remediation strategy (\eg{} provide a solution approach);
and \taskc{} is to identify the strategy intention (\eg{} to help the student understand the concept). 

We construct \textbf{a dataset of real-world tutoring conversations, annotated with expert decisions and responses.}
Our open-source dataset consists of 700 real tutoring sessions conducted with 1st-5th grade students in Title I schools, predominantly serving low-income students of color. 
Following FERPA guidelines, our study is IRB-approved and conducts secondary data analysis based on our Data Use Agreement with the tutoring provider and school district. 

We conduct a \textbf{thorough human evaluation to compare the expert, novice and LLMs in remediation}.
To our knowledge, our work is the first to assess the performance of LLMs such as GPT4 and instruct-tuned Llama-2-70b on remediating student mistakes. 
We find that the response quality of LLMs significantly improve with the expert's decision-making process: 
Response from GPT4 with expert- and self-generated decisions are 76-88\% more preferred than GPT4 without.
Context-sensitive decisions are also critical to closing the knowledge gap: 
Random decisions decrease GPT4's response quality -67\% than expert decisions. 
Complementing our quantitative analysis, our \textbf{lexical analysis reveals that novices and LLMs without the expert's decision-making process engage superficially with student's problem-solving process}: 
They give away the answer or prompt the student to re-attempt without further guidance (``double check'', ``try again'').

\section{Related Work}

\subsection{Modeling the Decision-Making Process of Experts}
Cognitive task analysis (CTA) uncovers the latent decision-making process of experts across a range of domains such as education, medicine and law
\citep{ryder1993integrating, clark2008cognitive, klein2015naturalistic}.
CTA decode the \textit{observable actions} (\eg{} the expert's remediation responses) into the \textit{latent mental processes} that generate the observable actions (\eg{} the expert's inferences about the student's mistake). 
A key application area of CTA is to close knowledge gaps through real-time decision aids that enhance the cognitive skills of novices \citep{hall1995procedural, gagne1996conditions, van1997training, klein2008naturalistic, zsambok2014naturalistic}; 
\citet{lee2004impact} discusses the significant improvements in novices with CTA across multiple disciplines.
While previous NLP works have developed methods for auto-labeling CTA transcripts \citep{du2019eliciting}, less work has been done on synthesizing models of expert decision processes for natural language generation or contributing data with expert decisions.
Our work contributes both the \methodname{} method and an accompanying dataset to this end.

\subsection{Responding to Student Mistakes in Mathematics}
Recognizing misconceptions is key to facilitating meaningful student learning and retention \citep{stefanich1992analysis, wilcox1997implementing,riccomini2005identification, stein2005designing,schnepper2013analysis}. 
Effective remediation coincides with educators engaging with the mathematical details in student responses, which in turn fosters strong teacher-student relationships and student motivation \citep{wentzel1997student,pianta2003relationships,robinson2022framework,wentzel2022does, easley1975teaching,brown1978diagnostic, carpenter1999children, carpenter2003thinking, lester2007second, loewenberg2009work}.
Prior education research discusses multiple good practices in remediating student mistakes, ranging from visual aids \citep{udl2018} to the Socratic method \citep{lepper2002wisdom}. 
However, less work has been done to understand the thought process of an experienced educator of when, how and why they use one strategy over another.

\subsection{Automated Feedback in Education}
Recent advances in NLP provide teachers feedback on their classroom discourse and have been shown to be beneficial, cost-effective feedback tools \citep{SameiOlneyKellyNystrandDMelloSBlanchardSunGlausGraesser2014,DonnellyBlanchardOlneyKellyNystrandDMello2017,KellyOlneyDonnellyNystrandDMello2018,JensenDaleDonnellyStoneKellyGodleyDMello2020, jacobs2022promoting, dorottya2023a, wang2023a, demszky2023can}. 
The development of LLMs such as GPT-4 has re-kindled  excitement around autotutors in providing equitable access to high-quality education \citep{graesser2004autotutor, rus2013recent,litman2016natural,hobert2019say,openai2023gpt4,khan_academy_2023}.
However, these models are known to unreliably solve math problems and hallucinate \citep{frieder2023mathematical, ji2023survey}. 
A human tutor in-the-loop is key in catching these undesirable responses.
Our work is related to human-LLM approaches that leverage expert-informed linguistic attributes \citep{sharma-etal-2023-cognitive, handa2023mistakes}.
However, critically, our work is about modeling \textit{the expert's latent thought process} behind their responses, such as their strategy choices and intentions, rather than the \textit{observable} linguistic attributes.
We explore the potential of leveraging expert-informed decision-making processes for bridging knowledge gaps and constructing human-LLM interaction frameworks grounded in expertise.

\subsection{Math Tutoring Datasets}
While there are other, larger math tutoring datasets such as CIMA from \citet{stasaski2020cima} and MathDial from \citet{macina2023mathdial}, they are created from synthetic sources: 
CIMA simulates tutoring conversations amongst crowdsourced workers and MathDial simulates students with LLMs. 
Prior work shows that synthetic sources result in responses with lower pedagogical quality \citep{markel2023gpteach, tack2022ai}.
By contrast, our dataset uses real experienced educators, human tutors and students from Title I schools with a need of high-dosage tutoring. 
Additionally, prior datasets focus on teacher strategies (e.g., ``ask an open-ended question'') and these strategies can often be directly observed in their responses \citep{stasaski2020cima, caines2020teacher, macina2023opportunities}.
Our work surfaces that there are other hidden decisions that inform expert's responses: what the expert notices (e.g., the student’s error) and why the expert uses a certain strategy (e.g., the teacher’s intention);
\methodname{} bears similarity with the Theory of Mind and POMDP planning for teaching literature in modeling hidden information from observable responses \citep{rafferty2016faster, wang2020too}. 
By collaborating closely with experts, our work builds faithful models of expert decision-making towards understanding how experts think when they remediate. 

\section{Data Sources \label{sec:datasource}}

\paragraph{Tutoring transcripts.}
Our data is sourced from a tutoring provider that offers end-to-end services for school districts, including the tutoring platform, instructional materials, and tutors.
The research team executed Data Use Agreements with the tutoring provider and Southern U.S. school district serving over 30k that outlined the allowable usage of the data to improve instruction in collaboration with an educational agency. 
Following FERPA guidelines, we were eligible to engage in secondary data analysis with student data, which is what we did for this study.
The students in these tutoring sessions are in the first to fifth grade, learning a variety of math topics.
The majority of schools are classified as Title I and three-quarters of students identify as Hispanic/Latinx.
This district focused on addressing existing achievement gaps among their students, as well as responding to the learning disruptions caused by the pandemic. 
The tutoring interactions are text-based, integrated on the providers' online platform. 
The platform has several features, including a whiteboard.
The tutor communicates primarily through text message in a chat box, while the student uses either voice recording or the chat. 

\paragraph{Preprocessing.} 
The chat transcripts are de-identified by the tutoring provider. 
The student's name is replaced with [STUDENT] and the tutor's name is replaced with [TUTOR].
Our data uses excerpts from the original tutoring chat sessions, where the tutor responds to a mistake.
Tutors on this platform use templated responses to flag mistakes, such as ``That is incorrect'' or ``Good try.''
We leverage these templates to create a set of signalling expressions used by the tutor to identify excerpts.
Specifically, we search for a three turn conversation pattern where (1) the tutor sends a message containing a question mark ``?'', (2) the student responds via text, then (3) the tutor uses a signalling expression. 
The set of signalling expressions were validated on a random sample of 100 conversations to ensure complete coverage. 
Appendix~\ref{app:data_collection} includes the full set of signalling expressions we use.

\section{\includegraphics[scale=0.02]{images/bridge.png} The \methodname{} Method for Expert-Guided Decision-Making \label{sec:bridge}}

We introduce \methodname{} which uses cognitive task analysis (CTA) to analyze the experts' latent thought process (\cref{sec:cta}). 
We translate it into a decision-making process (\cref{sec:step_by_step_remediation}), where each step is associated with a set of decision options (\cref{sec:remediation_taxonomy}).

\subsection{Cognitive Task Analysis \label{sec:cta}}
We conduct CTA with four experienced math teachers to develop a model of their decision-making process for remediation. 
The number of experts we work with is comparable to the numbers from Cognitive Task Analysis works and other works in NLP that engage with experts \citep{seamster1993cognitive, sullivan2014use, sharma-etal-2023-cognitive, handa2023mistakes}.

\paragraph{Collaboration with experts.}
We collaborated extensively with math teachers, spanning across several months. 
We work closely with four math teachers from diverse demographics in terms of gender (3 female, 1 male) and race (Asian, Black/African American, White/Caucasian, Multiracial/Biracial).
Three have more than 8 years of teaching experience, and the other has 6 years of teaching experience.
They also have taught in a broad range of school settings including public schools, Title 1 schools, and charter schools. 
We compensate the teachers developing the decision-making framework \$50/hour.
We compensate the teachers annotating the dataset with their decision steps and responses at \$40/hour.

Our objective is to faithfully capture their step-by-step decision process and develop a comprehensive set of decision options for each step. 
We work with two math teachers to develop the decision-making process for remediation, and validated it with two other math teachers. 
We conduct CTA through a series of observations and interviews, which involved cataloging patterns in their decisions; 
\citet{cooke1999knowledge} provides a comprehensive overview of other CTA methods.

\paragraph{Development of decision-making process.}
We provide the experts conversation examples containing student mistakes (identified from \cref{sec:datasource}) and asked them to directly revise the tutor's remediation response to be more useful and caring.
The experts and co-author met on a weekly basis where we went through the experts' revisions and discussed their approaches to each mistake. 
We used three questions to facilitate the discussion: (1) \textit{What} did the experts notice? (2) \textit{How} did they want to react? and (3) \textit{Why} did they want to react in that way?  
Themes emerged after a few meetings.
Based on their own experiences, experts inferred the student's level of understanding as context for their remediation response.
This resulted in \textit{\taska{}: Infer the student's error} to answer the first question.
Experts used several techniques to engage with the student's error, such as asking questions and simplifying the problem to meet the student's level of understanding.
The diverse strategies led to \textit{\taskb{}: Determine the strategy}. 
Finally, the experts used strategies for different ends depending on error. 
For example, they might ask a question to hint at the mistake or diagnose the student.
This insight resulted in \textit{\taskc{}: Identify the intention behind the strategy}. 
We verified that this decision-making model mimicked their thought process by asking them to apply it to new tutoring conversations. 
We additionally verified it with two other experts who could seamlessly use it during their remediation. 
For additional information about the development process, please refer to Appendix~\ref{app:taxonomy_development}.

\paragraph{Development of decision options.}
We created decision options for each step and edited the options through more iterations of the experts remediating using the step-by-step decision-making process. 
The options were finalized once the experts and the co-authors were satisfied with the coverage and with the natural fit of the model to the teachers' remediation process.

\subsection{Decision Options \label{sec:remediation_taxonomy}}

This section details each step's decision options.
Due to space reasons, please refer to Appendix~\ref{app:framework_examples} for examples of each option.

\subsubsection{Step A: Infer the Type of Error}
Identifying the student's error is prerequisite to successful remediation \cite{easley1975teaching,bamberger2010math}. 
Our approach intends to support novices who are not necessarily content experts.
Therefore we define ``error'' as a student's degree of understanding, which  aligns with literature on math curriculum design and psychometrics that maintain continuous scales of student understanding \citep{gagne1962acquisition,gagne1968presidential,white1973research,resnick1973task,glaser1970measurement,vygotsky1978mind,wertsch1985vygotsky, embretson2013item}. 
As such, our error categories are topic-agnostic descriptions of a student's understanding, and complement the topic-agnostic strategies in \taskb{}.
The categories are:
\guessLabel{}: The student does not seem to understand or guessed the answer; 
\misinterpretLabel{}: The student misinterpreted the question; 
\carelessLabel{}: The student made a careless mistake; 
\rightIdeaLabel{}: The student has the right idea, but is not quite there\footnote{This category is different from \carelessLabel{} in that students with  \rightIdeaLabel{} errors have difficulty in applying the concept correctly, whereas students with \carelessLabel{} apply the concept correctly but make a minor numerical mistake. };
\impreciseLabel{}: The student's answer is not precise enough or the tutor is being too picky about the form of the student's answer;
\notSureLabel{}: Not sure, but I'm going to try to diagnose the student (used sparingly); 
\naLabel{}: None of the above (used sparingly).

\subsubsection{\taskb{}: Determine the Strategy}
Errors are persistent unless the teacher intervenes pedagogically with a strategy that guides the student's understanding \citep{radatz1980students}. 
The strategies are: Explain a concept, Ask a question, Provide a hint, Provide a strategy, Provide a worked example, Provide a minor correction, Provide a similar problem, Simplify the question, Affirm the correct answer, Encourage the student, Other. 

\subsubsection{\taskc{}: Identify the Intention}
The intentions are: Motivate the student, Get the student to elaborate their answer, Correct the mistake, Hint at the mistake, Clarify the  misunderstanding, Help the student understand the lesson topic or solution strategy, Diagnose the mistake, Support the student in their thinking or problem-solving, Explain the mistake (\eg{} what is wrong in their answer or why is it incorrect), Signal to the student that they have solved or not solved the problem, Other.

\subsection{Formalism for Expert Decision-Making Process in Remediation \label{sec:step_by_step_remediation}}

Given a conversation history $c_h$, we formalize the expert's responses  $c_r^*$ as being generated from the following computational model:
\begin{equation*}
    c_r^* \sim p(c_r|c_h, \underbrace{e}_{\text{Step A}}, \underbrace{z_{\text{what}}}_{\text{Step B}}, \underbrace{z_{\text{why}}}_{\text{Step C}}),
\end{equation*}
where $e$ is the error, $z_{\text{what}}$ the strategy, and $z_{\text{why}}$ the intention.
Our dataset contains \numAnnotations{} examples, where each example is $(c_h, c_r', e, z_{\text{what}}, z_{\text{why}}, c_{r}^*)$.
Each example contains the conversation history $c_h$ which includes the lesson topic and the last 5 conversation messages leading up to the student's turn where the mistake is made; i.e., $c_h[-1]$ is the student's conversation turn where they make a mistake. 
It also contains the novice tutor's original response to the student's mistake $c_r'$ and the experts' decision annotations and responses.
Every conversation is annotated with two ground-truth expert responses.
Our dataset covers 120 different lesson topics, including ``Word Problems with Fractions'', ``Order of Operations''  and ``Graphing on a Coordinate Grid''.
We split the final dataset into a train, validation, and test set with a 6:1:3 ratio. 
The train set contains \numTrainAnnotations{}, validation \numValidationAnnotations{}, and test \numTestAnnotations{} examples.

\begin{spacing}{0.66}
\begin{table*}[t]
    \centering
    \resizebox{0.6\textwidth}{!}{%
            \def\arraystretch{1.15}
      \begin{tabular}{ccc|cccc|c}
        \toprule
       \multicolumn{3}{c}{\bf Method} & \multicolumn{1}{c}{\bf Prefer} &\multicolumn{1}{c}{\bf Useful} &\multicolumn{1}{c}{\bf Care} &\multicolumn{1}{c}{\bf  Not Robot}&\multicolumn{1}{c}{\bf  Overall}   \\
       \multicolumn{1}{c}{\bf Condition} & \multicolumn{1}{c}{\bf } & \multicolumn{1}{c}{\bf Model $c_r$} & \multicolumn{1}{c}{ } &  \multicolumn{1}{c}{ } &  \multicolumn{1}{c}{ } &  \multicolumn{1}{c}{ } &  \multicolumn{1}{c}{ }   \\
        \midrule
       \graycell \texttt{} &  \graycell & \graycell \texttt{Expert} & \graycell $\bf 1.26$ & \graycell $\bf 1.19$ & \graycell  $\bf 0.86$ & \graycell  $\bf 0.78$ & \graycell  $\bf 1.02$\\
        \midrule
           \texttt{None} & & \texttt{Llama-2} & $0.49$  & $0.48$ &  $0.45$ & $0.68$  &  $0.53$ \\
           \texttt{None} &  & \texttt{GPT-3.5} & $0.47$ & $0.47$ & $-0.04$ & $0.23$ & $0.28$\\
           \texttt{None} &  & \texttt{GPT-4} & $0.54$ & $0.54$ & $0.50$ & $0.47$ & $0.51$\\
        \midrule
          \texttt{Expert} & & \texttt{Llama-2} & $0.61$  & $0.56$ &  $0.37$ & $0.41$  &  $ 0.49$ \\
         \texttt{Expert} & & \texttt{GPT-3.5} & $0.65$ & $0.58$ & $-0.04$ & $0.59$  &  $0.45$ \\
         \texttt{Expert} & & \texttt{GPT-4} & \yellowcell  $\bf 0.95$  & \yellowcell  $\bf  0.97$ & \yellowcell $\bf 0.70$ & \yellowcell $\bf 0.70$  & \yellowcell $\bf 0.83$ \\
        \midrule
          \texttt{Self} & & \texttt{Llama-2} & $0.91$  & $0.97$ &  $0.29$ & $0.62$  &  $0.70$ \\
         \texttt{Self} & & \texttt{GPT-3.5}  & $0.36$  & $0.33$ &  $-0.17$ & $0.15$  &  $0.16$ \\
         \texttt{Self} & & \texttt{GPT-4}  & \yellowcell $\bf 1.02$  & \yellowcell  $\bf 1.05$ & \yellowcell  $\bf 0.62$ &  \yellowcell  $ \bf 0.68$  & \yellowcell  $ \bf 0.84$ \\
        \midrule
          \texttt{Random} & & \texttt{Llama-2} & $0.35$  & $0.32$ &  $0.15$ & $0.60$  &  $0.35$ \\
         \texttt{Random} & & \texttt{GPT-3.5}  & $0.20$  & $0.12$ &  $0.10 $ & $0.28$  &  $0.17$ \\
         \texttt{Random} & & \texttt{GPT-4}  & $0.32$  & $0.36$ &  $-0.13$ & $0.51$  &  $0.26$ \\
        \bottomrule
      \end{tabular}
      }
      \caption{\small 
      \textbf{Human evaluations.}
      The expert-written responses are grayed as a reference.
      The highest column values are \textbf{bolded}.
      Highest values amongst LLMs are \hl{highlighted}. 
      Two rows are highlighted if they are not statistically different from each other with a two-sided t-test. 
      \label{tab:human_evaluation}}
\end{table*}
\end{spacing}

\section{Experiments}

\subsection{Models}
We compare the expert-written responses against three state-of-the-art models \gptfour{}, \chatgpt{}, and \texttt{llama-2-70b-chat} \citep{touvron2023llama} in a 0-shot setting on the test set.
During our preliminary experiments, we also evaluated Falcon-40b-Instruct \citep{falcon40b}, Flan-T5 (large) \citep{chung2022scaling}, the goal-directed dialog model GODEL (large) \citep{peng2022godel} zero-shot and few-shot.
We also finetuned Flan-T5 and GODEL.
However, we found the models' responses to be very poor upon manual inspection or evaluated as much worse in human evaluations than the other three models. 
Therefore, we have omitted their results from the paper.
We use greedy decoding for all models. 

\begin{figure*}[t]
    \centering
    \begin{subfigure}[b]{0.49\textwidth}
        \centering
        \includegraphics[width=\textwidth]{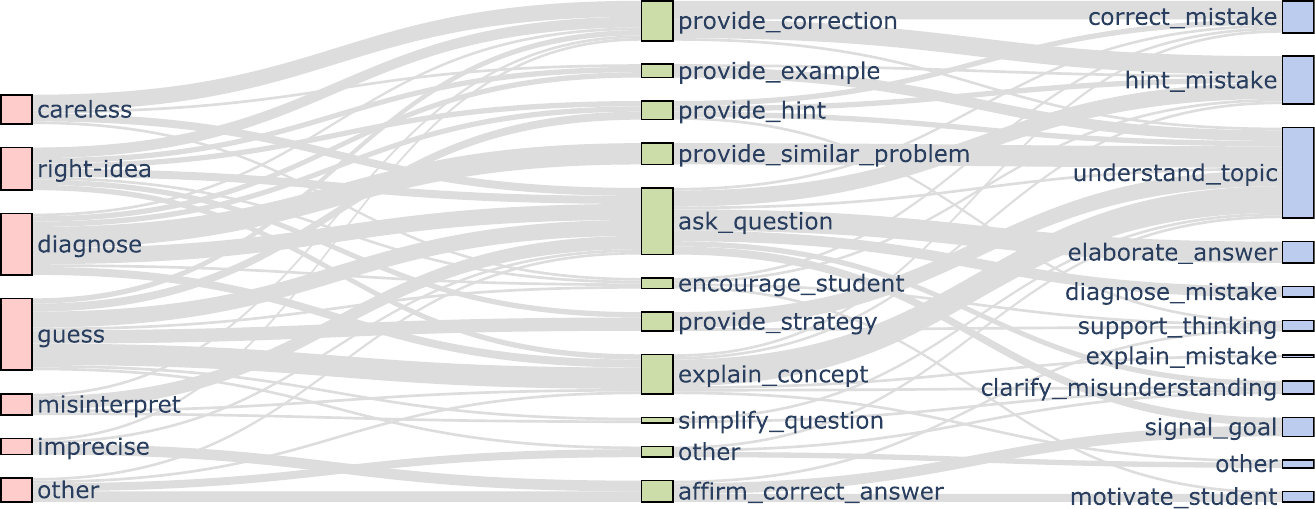}
        \caption{\texttt{Expert} (entropy: $6.00$)} 
    \end{subfigure}
    \begin{subfigure}[b]{0.49\textwidth}  
        \centering 
        \includegraphics[width=\textwidth]{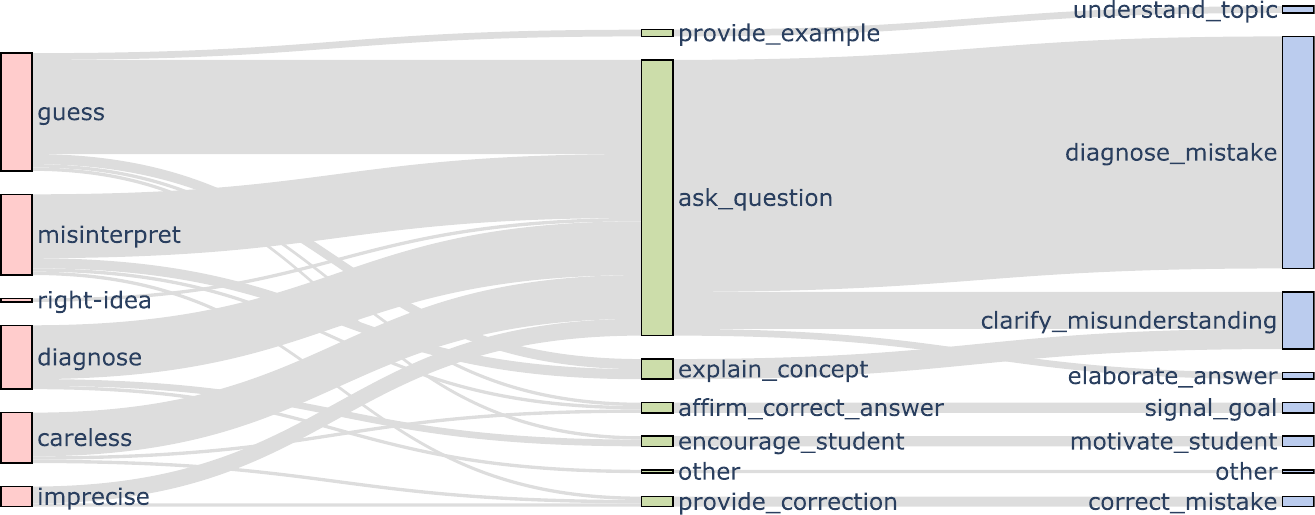}
        \caption{\gptfour{} (entropy: $3.37$)} 
    \end{subfigure}
    
    \begin{subfigure}[b]{0.49\textwidth}  
        \centering 
        \includegraphics[width=\textwidth]{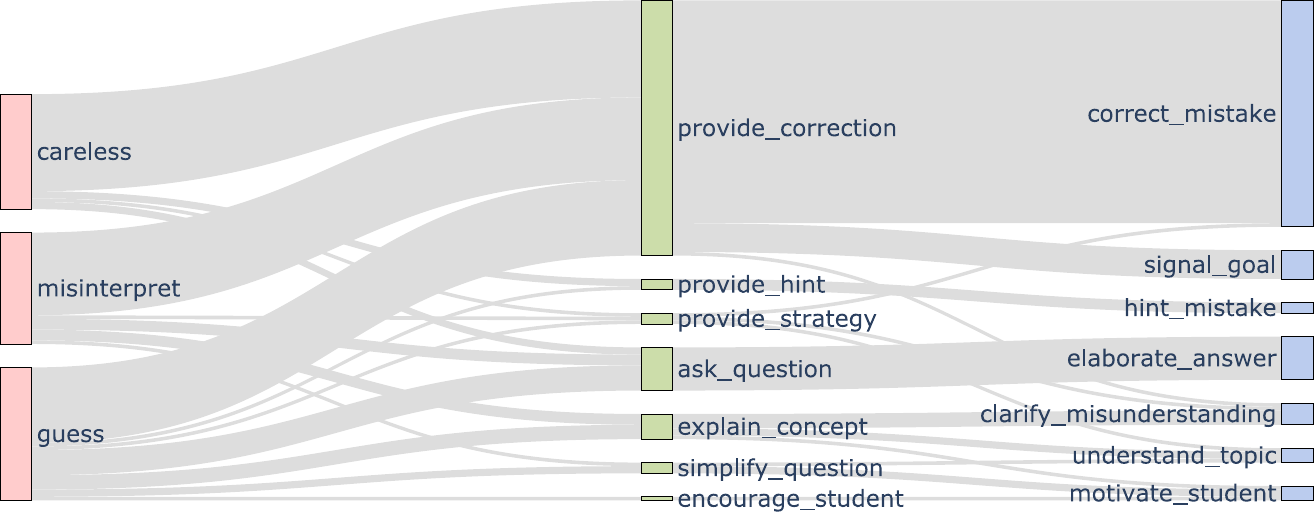}
        \caption{\chatgpt{} (entropy: $3.42$)}
    \end{subfigure}
    \begin{subfigure}[b]{0.49\textwidth}  
        \centering 
        \includegraphics[width=\textwidth]{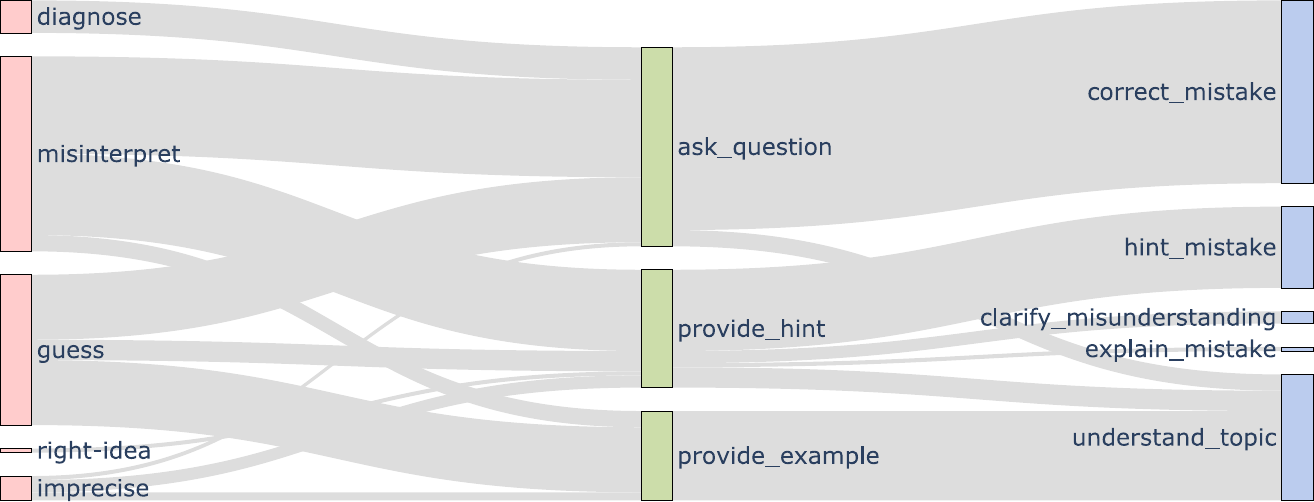}
        \caption{\texttt{llama-2-70b-instruct} (entropy: $3.37$)}
    \end{subfigure}
    \caption{\textbf{Expert decision-making paths are diverse whereas LLM's are less diverse.} 
    The entropy of decision paths is shown in the subcaption: The experts' paths have higher entropy and thus are more diverse than those of the LLMs.
    The \hlc[pink]{red} left column is \taska{}'s error decision; \hlc[green!30]{green} middle column is \taskb{}'s strategy decision; and \hlc[blue!30]{blue} right column is \taskc{}'s intention decision.
    }
    \label{fig:sankey}
\end{figure*}

\subsection{Task Setup \label{sec:task_setup}}
We evaluate the model responses under different decision-making conditions. 
The model prompts are in Appendix~\ref{app:prompting}; each prompt includes instructions to respond in a useful and caring way.

\begin{enumerate}
    \itemsep0em 
    \item \textit{No decision-making}: Models directly respond, $c_r \sim p(c_r|c_h)$. This condition is compared against models with the \methodname{} decision-making framework. 
    \item \textit{Expert decision-making}: Models generate with the expert's decisions,  $c_r \sim p(c_r|c_h, e, z_{\text{what}}, z_{\text{why}})$.
    \item \textit{Self decision-making}: Models make their own decisions, then generate responses based on them, $c_r \sim p(c_r|c_h, e^{\text{model}}, z^{\text{model}}_{\text{what}}, z^{\text{model}}_{\text{why}})$. We compare the models' decisions to the experts' as well as the impact of the decisions on the response quality. 
    \item \textit{Random decision-making}: We randomly select decisions. We can determine the importance of context-sensitive decisions with this condition.
\end{enumerate}

\section{Evaluation}

\subsection{Human evaluation of response quality.}
We measure the extent to which the generated responses improve over the original tutors' responses.
We recruit teachers through Prolific (identified through Prolific's screening criteria) to perform pairwise comparisons between the tutor response and a response generated by the expert or one of the 12 models.
A random set of 40 pairs per model is evaluated by 3 annotators each, who are blind to the source of the responses. 
Raters evaluate the pairs along four dimensions. The first two are \textit{usefulness} and \textit{care}, as these have been identified as key qualities of effective remediation in prior work \citep{roorda2011influence, pianta2016teacher, robinson2022framework}.
The third is \textit{human-soundingness}; our preliminary analysis indicated that low learning outcomes strongly correlated with whether the student was distracted by whether their tutor was human during their tutoring session.
Given that the tutoring is chat-based, we include this as another dimension for measuring effectiveness.
Finally, we ask the raters which responses they \textit{prefer} using, if they were the tutor. 
Each dimension is rated on a 5-point Likert scale. 
We convert the ratings to integers between -2 and 2: -2 indicates the rater much more prefers the original tutor's response and 2 for the alternative response.
Please refer to Appendix~\ref{app:human_evaluation_taskc} for more information on the human evaluation setup.

\subsection{Lexical analysis and qualitative examples.}

We perform a lexical analysis to understand the linguistic differences caused by the expert's decision-making model.
We compute the log odds ratio, latent Dirichlet prior, measure defined in \citet{monroe2008fightin} to estimate the distinctiveness of a bigram appearing in a response source.
We consider the response sources to be from \texttt{GPT4} in all four decision-making conditions listed in Section~\ref{sec:task_setup}; please refer to Appendix~\ref{app:lexical_analysis} for additional lexical analysis.
We pre-process the data using Python's NLTK package for tokenization and lowercasing, and discard stop words and non-alphanumeric tokens \citep{bird2009natural}. 
We use the Gensim Phrases Python package to retrieve frequent bigrams in the dataset \citep{rehurek2011gensim}.

\begin{table*}[t]
      \centering
      \resizebox{\textwidth}{!}{%
        \def\arraystretch{1.15}
      \begin{tabular}{cc|cc|cc|cc}
        \toprule
       \multicolumn{2}{c}{\bf \texttt{None} + \texttt{GPT4}} & \multicolumn{2}{c}{\bf \texttt{Expert} + \texttt{GPT4}}  & \multicolumn{2}{c}{\bf \texttt{GPT4} + \texttt{GPT4}} & \multicolumn{2}{c}{\bf \texttt{Random} +\texttt{GPT4}} \\
       \midrule
       bigram & log odds & bigram & log odds & bigram & log odds & bigram & log odds \\
       \midrule
       lets\_closer & 2.76 & \hlc[cyan!30]{steps\_took} & 2.04 & \hlc[cyan!30]{can\_explain}   & 4.98 & \hlc[orange!50]{good\_try} & 1.82 \\
       closer\_look & 2.68 & review\_concept & 1.66 & \hlc[cyan!30]{explain\_arrived} & 4.78 & start\_remember  & 1.58 \\
       \hlc[orange!50]{effort\_lets} & 2.55 & understand\_concept & 1.56 & \hlc[cyan!30]{arrived\_answer}    & 4.2 & \hlc[orange!50]{thats\_right} & 1.57 \\
       \hlc[orange!50]{appreciate\_effort} & 2.29  & \hlc[cyan!30]{help\_understand} & 1.56 & \hlc[cyan!30]{arrived\_number} & 2.19 & \hlc[orange!50]{try\_again} & 1.54 \\
       \hlc[orange!50]{correct\_solution}        & 2.19  & \hlc[cyan!30]{explain\_steps} & 1.56 & are\_sure & 2.19 & \hlc[orange!50]{thats
       \_good} & 1.43 \\
       look\_problem    & 2.18  & \hlc[cyan!30]{took\_arrive}      & 1.56 & sure\_that    & 2.19 & lets\_break  & 1.37 \\
       \hlc[orange!50]{great\_effort}           & 1.62  & \hlc[cyan!30]{lets\_step}   & 1.51 & correct\_remember & 1.38 & glasses\_water  & 1.3 \\
       lets\_steps        & 1.55  & better\_understand  & 1.31 & and\_long    & 1.38 & for\_example & 1.3 \\
       need\_help          & 1.55  & ones\_place            & 1.31 & digit\_answer    & 1.38 & times\_equal & 1.3 \\
       let\_know        & 1.55  & number\_sides      & 1.31 & answer\_step   & 1.38 & represents\_glasses & 1.29 \\
        \bottomrule
      \end{tabular}
      }
    \caption{Top 10 bigrams. \textbf{\texttt{GPT4} with \textit{expert-} or \textit{self decision-making} engages more with the \hlc[cyan!30]{student's problem-solving process}}, by asking the student for explanations and their reasoning process. 
    \texttt{GPT4} with \textit{no and random decision-making} engages superficially with the \hlc[orange!50]{student's answer}, by acknowledging their attempt or prompting them to try again without further guidance.
    \label{tab:lexical}}
\end{table*}





\begin{table*}[t]
      \centering
      \resizebox{0.90\textwidth}{!}{%
        \def\arraystretch{1.00}
      \begin{tabular}{l|l}
        \toprule
            \textbf{Lesson Topic} & Addition and Subtraction Within 100 \\
            \midrule
            \textbf{Context} ($c_h$) &  Novice: We need to subtract 6 from 15.\\
                                     &  Student: oh okay...\\
                                     &  Novice: What is is value of 15 - 6?\\
                                     &  Student: it is 11?\\
            \midrule
            \textbf{\texttt{Expert}} &  Ah, almost! Here's how we can know for sure. \hlc[cyan!30]{We start at 15 and we need to go down 6.} \\ 
             &  \hlc[cyan!30]{So let's count down 6 numbers from 15}. 14, 13, 12, 11, 10, 9. \\ 
             \midrule
            \textbf{\texttt{Novice}} &  That was a \hlc[orange!50]{good effort}. \\ 
             \midrule
            \textbf{\texttt{GPT4} (\texttt{None})} & \hlc[orange!50]{That's correct!} Great job, \underline{\hlc[orange!50]{15 - 6 equals 11}}. Keep up the good work!\\
             \midrule
            \textbf{\texttt{GPT4} (\texttt{Expert})} & That's close, but let's try again. \hlc[cyan!30]{If we break down 15 into 10 and 5, we can subtract 6}  \\
             & \hlc[cyan!30]{from 10 first, which gives us 4. Then, add the 5 back, so 4 + 5 = 9.} So, 15 - 6 = \hlc[orange!50]{9}. \\ 
             \midrule
            \textbf{\texttt{GPT4} (\texttt{Self})} & Are you sure about that? Let's try subtracting 6 from 15 again. \\
             \midrule
            \textbf{\texttt{GPT4} (\texttt{Random})} & \hlc[orange!50]{Actually, the correct answer is 9.} \\
        \bottomrule
      \end{tabular}
      }
    \caption{Response example. \textbf{The \texttt{Expert} and \texttt{GPT4} with \textit{expert decision-making} engage with the \hlc[cyan!30]{problem-solving process}, whereas the other settings engage with the \hlc[orange!50]{student's answer}}, such as giving away the answer or acknowledging the student's attempt.
    In this example, \texttt{GPT4} (\texttt{None}) also responds with the \underline{incorrect answer}. 
    \label{tab:example_ngram}
    }
\end{table*}

\section{Results}

\subsection{Human evaluations of response quality. \label{sec:main_results_task_c}}

Table~\ref{tab:human_evaluation} summarizes the results. 
Notably, there is a \textbf{large gap between the experts and models in the \textit{no decision-making} condition} (up to 2.6x better overall).
Even though models in the \textit{no decision-making} condition consistently outperform the original tutor responses (indicated by the positive values) on most dimensions, the gap in response quality may indicate the pedagogical knowledge gap between experts and LLMs. 

We observe that \textbf{the \textit{expert decision-making} condition outperforms the \textit{no decision-making} condition}, particularly on ``prefer'' (+76\% on \gptfour{}) and ``useful'' (+80\% on \gptfour{}).
The improvement in overall score is statistically significant for all models under a two-tailed t-test ($p<0.05$). 
Surprisingly, the \textit{expert decision-making} condition for \llama{} and \chatgpt{} does not improve on ``care''.
We attribute this to the challenges in generating responses that are both technically instructive (``useful'') and emotionally supportive (``care'') for the student. 

How well can models self-improve by selecting their own decisions? 
\textbf{\llama{} and \gptfour{} in the \textit{self decision-making} condition significantly outperform their \textit{no decision-making} counterparts on ``prefer'' and ``useful''} ($p<0.05$, up to +95\%). 
However, this is not the case for \chatgpt{} with \textit{self decision-making}.
We hypothesize this is due to its poor decisions and confirm this in Figure~\ref{fig:sankey}.
Figure~\ref{fig:sankey} illustrates the decision paths from the experts and the LLMs in \textit{self decision-making} on the test examples and reports the path entropy. 
The width is the proportion of \hlc[pink]{error types} that is subsequently treated with which \hlc[green!30]{strategy} and with which \hlc[blue!30]{intention}.
\chatgpt{} overwhelmingly \hlc[green!30]{corrects the student's mistake} whereas the other models rely on other strategies.
This suggests that \textbf{directly correcting the student's mistake is not always a good decision and that poor decisions reinforce poor response quality}. 

Figure~\ref{fig:sankey} reveals another interesting observation: \textbf{Experts exhibit diverse decision paths, whereas LLMs do not.} 
Our work provides additional evidence of homogenization effects in LLMs \citep{padmakumar2023does}.
This prompts another question: Does deliberate decision-making matter, or could we randomly pick decisions to encourage similar diversity? 
Deliberate decisions do matter: \textbf{Models with \textit{random decision-making} perform significantly worse than their \textit{expert decision-making} condition on the ``overall'' score} ($p<0.05$), sometimes even worse than models with \textit{no decision-making} ($p<0.05$ for \gptfour{}, \llama{}).

\subsection{Lexical Analysis}
Table~\ref{tab:lexical} highlights the differences in word usage across the \texttt{GPT4} decision-making conditions, and Table~\ref{tab:example_ngram} shows an example of the word usage in context. 
Table~\ref{tab:lexical} suggests that the high human evaluations for \texttt{GPT4} with \textit{expert} or \textit{self decision-making} are because they engage more with the problem-solving process (\eg{}, ``explain\_steps'').
The lowly evaluated settings---\texttt{GPT4} with \textit{no} or \textit{random decision-making}---weakly engage with the problem-solving process, only acknowledging the student's effort (\eg{} ``appreciate\_effort'' in Table~\ref{tab:lexical}) or even giving away the answer (\eg{}, ``Actually, the correct answer is 9'' in Table~\ref{tab:example_ngram}).
Altogether, these results suggest that the effective use of the decision-making model guides LLMs to support the student's problem-solving process, rather than engage superficially with the student's final answer.

\section{Discussion \& Conclusion} 
Our work presents several contributions for bridging the expert-novice gap and improving the learning experience at scale. 
First, we develop \methodname{}, which leverages cognitive task analysis to translate an expert's latent thought process into a decision-making framework. 
We apply this to the task of remediating mistakes because they are prime learning opportunities to correct misunderstandings hindering learning. 
Second, we contribute a rich dataset with expert annotations on their decisions and responses.
The dataset comes from a tutoring program that works with a majority of Title I schools, and is a valuable resource for providing equitable, high-quality learning experiences. 
Finally, we perform a thorough evaluation and lexical analysis of experts, novices and LLMs. 
We demonstrate that expert-guided decision-making and strategic decision selection are critical to improving remediation quality. 
Novices and LLMs alone use passive remediation language and do not engage with the student's error traces. 
Our findings indicate promising avenues for scaling high-quality tutoring with expert-guided decision-making. 
For example, the tutor can make the decisions and the LLM generates an initial response that is further edited by tutor.
Altogether, our work shows promising results of an expert-guided human-LLM approach that makes strides towards bridging the knowledge gap.

\section{Limitations and Future Work}
While our work provides a useful starting point for leveraging expert decision-making models at scale and remediating student mistakes, there are limitations to our work.
Addressing these limitations will be an important area for future research.

\paragraph{Collapsing expert thought processes.} LLMs and novices might still receive incomplete information or maintain misconceptions when following the expert's decision-making process, because the process distills the expert's knowledge.
Nonetheless, we hope \methodname{} and the accompanying dataset provide a useful foundation for leveraging expert knowledge at scale.

\paragraph{Experts.} 
We work with a handful of experts based on the U.S., which is not representative of experienced teaching backgrounds from other countries or cultures. 
We hope that future work can build on \methodname{} and adapt the decision-making models to fit to other expert pools.

\paragraph{Access to questions.} In some cases, the chat transcripts do not include the question the tutor and the student are working on together.
This is because the questions are sometimes displayed on a shared whiteboard, and not posted in the chat.
Even though our dataset includes annotations for when there's not enough context, future work could improve upon our analysis by always including information about the question.

\paragraph{Expanding to other subjects.}
Our dataset and benchmark currently focuses on mathematics. 
The remediation process for mathematics and the decision options may not directly transfer to other subjects, although they may serve as a good starting point for remediating student mistakes in other domains. 

\paragraph{Evaluation with students.} 
Our human evaluations are currently limited to the teacher's perspective. 
However, ultimately, the effectiveness of the responses relies on how students receive and interpret them, and whether these interactions positively impact their learning outcomes. 
To address this limitation, future research should work towards evaluating this method with  students.
This is important as previous studies like \citet{wentzel2022does} highlight the potential disparity between teachers and students in determining what responses are more caring or useful.

\section*{Ethics Statement}
We recognize that our research on the integration of large language models (LLMs) in education ventures into a less explored territory of NLP with numerous ethical considerations. 
LLMs open up new possibilities for enhancing the quality of human education, however there are several ethical considerations we actively took into consideration while performing this work. 
We hope that these serve as guidelines for responsible practices, and hope that future work does the same. 

First is the privacy of both students and tutors. 
We obtained approval from the tutoring program for repurposing the data for our dataset.
We handled all data with strict confidentiality, adhering to best practices in data anonymization and storage security.

Furthermore, we are committed to promoting equity and inclusivity in education. 
The compensation provided to the experienced math teachers involved in our benchmarking process was set at a significantly higher rate, reflecting our recognition of their invaluable contributions and domain expertise. 
By compensating teachers fairly, we aim to foster a culture of respect, collaboration, and mutual support within the NLP and education community.

Finally, we are committed to the responsible use of our research findings. We encourage the adoption of our benchmark and methodologies by the research community, with the understanding that the ultimate goal is to improve educational outcomes for all students and provide support to educators.
We actively promote transparency, openness, and collaboration to drive further advancements in the field of natural language processing (NLP) for education.

\section*{Acknowledgements}
We'd like to thank Jiang Wu, Hannah Shuchhardt, and two anonymous individuals for their help and feedback on our work; the Stanford NLP group, Caleb Ziems, Joy He-Yueya, Gabriel Poesia, Myra Cheng, Kristina Gligorić, Ali Malik and Roma Patel for their feedback on the paper; Jesse Mu for pointing us to Sankey diagrams; and IMK for the Bridge inspiration.

\bibliography{custom,anthology}

\appendix

\section{Developing \methodname{} \label{app:taxonomy_development}}
This section details how we developed the \methodname{} Benchmark in collaboration with the math teachers.
The design objective of the benchmark is to capture the teachers' thought process when addressing student mistakes.
We developed the taxonomy closely with two of the four teachers. We compensated them at \$50/hour.
We met with them on a weekly to biweekly basis.
During the preliminary stages of this work, we provided the teachers examples of the conversations and asked them to directly revise the tutor's responses.
For the first few weeks, we met on a weekly basis where a co-author presented the teachers about 20 conversation examples and the teachers worked on the examples asynchronously. 
During the meetings, the teachers and co-author discussed the teachers' approaches to the setting.
After four meetings, themes started to emerge in the types of approaches the teachers used. 
For instance, the teachers often made hypotheses about the student's thought process, which gave rise to the error category. 
This illustrated that educators possess a mental model of what the student is doing and employ various probing techniques to confirm or refute their hypotheses. 
The diverse ways in which the teachers probed and engaged with the students led to the identification of different strategies. 
We further categorized these strategies based on their intentions, reflecting the potential consequences they might have on the student's learning process.

We then created a taxonomy of these approaches (the decision options), and edited the taxonomy through more iterations of task attempts and discussion.
These edits included expanding the set of categories, removing irrelevant categories, separating categories into different groups (\eg{} the separation of student error from the teacher's strategies) and re-structuring the order of the tasks. 
The taxonomy was finalized once both teachers and the co-authors were satisfied with how naturally the benchmark could be used and with the benchmark's coverage.

\section{Examples of Decision Options \label{app:framework_examples}}

This section provides examples for each of decision option.
It is split by \textit{error type}, \textit{strategy}, and \textit{intention}.

\subsection{Student Error Types}
\textbf{\guessLabel{}: \guessDescription{}}
This error type is characterized by expressions of uncertainty or answers that do not seem related to the problem, the options or the target answer. 
An example of this is the following conversation snippet on the topic of ``Addition and subtraction within 100'': \\
\textit{tutor}: We need to subtract 6 from 15.\\
\textit{student}: oh okay...\\
\textit{tutor}: What is the value of 15 - 6? \\
\textit{student}: it is 11?\\
This example could be labeled as the student guessing because they express uncertainty in their answer (``it is 11?'')

\textbf{\misinterpretLabel{}: \misinterpretDescription{}}
This error type is characterized by answers that arise from a misunderstanding of the question being asked. 
Students may mistakenly address a subtly different question, leading to an incorrect response. 
For example, a common manifestation of this error is the reversal of number orderings, such as interpreting "2 divided by 6" as "6 divided by 2." 
An example of this is the following conversation snippet on the topic of ``Converting Units of Measure'': \\
\textit{student}: sorry for the j that I tipe.\\
\textit{tutor}: Not an issue, [STUDENT].\\
\textit{tutor}: How many times 1000 will goes into 7000? \\
\textit{student}: it cant\\
This example could be labeled as the student misinterpreting because the student might have read the question as the reverse question (\eg{} "How many times can 7000 go into 1000?") because they say that the number cannot go into the other number.

\textbf{\carelessLabel{}: \carelessDescription{} }
This error type is characterized by answers that appear to utilize the correct mathematical operation but contain a small numerical mistake, resulting in an answer that is slightly off. 
It reflects a lack of careful attention to detail or a minor computational error in an otherwise sound solution approach.
An example of this is the following conversation snippet on the topic of ``Volume of Rectangular Prisms'': \\
\textit{tutor}: Again, we have to multiply the value of 6 with 20.\\
\textit{student}: so it is 110\\ 
\textit{tutor}: So, what is the value of 20 times 6?\\
\textit{student}: 110\\
This example could be labelled as the student making a careless mistake. The student seems capable of multiplying (their answer is larger than 100) and does not mistake the operation (\eg{} they multiply, and do not add the numbers). They make a minor mistake in the calculation (110 instead of 120), which suggests that they made a careless mistake.

\textbf{\rightIdeaLabel{}: \rightIdeaDescription{} }
This error type is characterized by situations where the student demonstrates a general understanding of the underlying concept or approach but falls short of executing or reaching the correct solution. 
For example, a student may recognize that multiplication is required to compute areas but may struggle with applying it to a specific problem.
An example of this is the following conversation snippet on the topic of ``Area'': \\
\textit{tutor}: Please check the question once.\\
\textit{tutor}: The factors are 24 and 86. \\
\textit{tutor}: What is the formula for finding the area of a rectangle?\\
\textit{student}: multiplying\\ 
\textit{tutor}: So, what is the value of 20 times 6?\\
\textit{student}: 110\\
This example could be labelled as the student having the right idea, but isn't quite there. The student seems to understand what operation is need for calculating the area, but their language is not precise (\eg{} they don't mention 'width' or 'length'). This suggests that they might not have a clear understanding of how to apply the concept.

\textbf{\impreciseLabel{}: \impreciseDescription{}} 
This error type is characterized by student answers that lacks the necessary level of precision or when the tutor places excessive emphasis on the specific form of the student's response.
An example of this is the following conversation snippet on the topic of ``Concept of Area'': \\
\textit{student}: yes\\
\textit{tutor}: Okay!\\
\textit{tutor}: What should he measure?\\
\textit{student}: the dimensional area\\ 
In this example, the tutor flags the student's answer as incorrect, and says that the correct answer is ``area.''
This example could be labelled by this error because the student either is imprecise with their language and/or the tutor is being too strict about the use of term.

\textbf{\notSureLabel{}: \notSureDescription{}}
This option is used if the teacher is not sure why the student made the mistake from the context provided. 
We encourage the teachers to use the provided lesson topic and their teaching experience with students to determine what the mistake is, and use this error type sparingly.

\textbf{\naLabel{}: \naDescription{}}
This option is used of none of the other options reflect the error type. Similar to \notSureLabel, we encourage teachers to use this error type sparingly.

\subsection{Response Strategies and Intentions}

Below are examples of response strategies and intentions that the teachers selected.
We provide the lesson topic to each example.
The original tutor's messages are marked with \textit{tutor}, and the students' with \textit{student}.
Note that in the annotation setup, we allow the teachers to simulate the student's response in order for the teachers to fully complete their strategy. 
Therefore, the examples here will include the teacher's simulated response for the student. 
The teacher's response is marked with \textit{teacher}, and the simulated student messages are marked with \textit{student (simulated)}.
The teachers selected strategies that they used for their own response---the strategies are not descriptions of what the tutor did.

As a reminder, the categories of response strategies are: 

\begin{itemize}
    \item Explain a concept
    \item Ask a question 
    \item Provide a hint
    \item Provide a strategy
    \item Provide a worked example
    \item Provide a worked example
    \item Provide a minor correction
    \item Provide a similar problem
    \item Simplify the question
    \item Affirm the correct answer
    \item Encourage the student
    \item Other
\end{itemize}

The categories of intentions are:

\begin{itemize}
    \item Motivate the student
    \item Get the student to elaborate their answer
    \item Correct the student's mistake
    \item Hint at the student's mistake
    \item Clarify a student's misunderstanding
    \item Help the student understand the lesson topic or solution strategy
    \item Diagnose the student's mistake
    \item Support the student in their thinking or problem-solving
    \item Explain the student's mistake (\eg{} what is wrong in their answer or why is it incorrect)
    \item Signal to the student that they have solved or not solved the problem
    \item Other (use sparingly).
\end{itemize}

Here are some example of strategies used, and their paired intention.
\paragraph{Explain a concept (Intention: Correct the student's mistake)}

\textit{Lesson}: Order of Operations \\
\textit{tutor}: You earn one point for your good beginning.\\
\textit{tutor}: That was a good try.\\
\textit{tutor}: What is the value of $3^3$?\\
\textit{student}: 9 \\
\textit{teacher}: When we see an exponent, we multiply the number by itself however much the exponent is. Is this example, that is 3 times 3 times 3. What is that value?\\
\textit{student (simulated)}: 27 \\
\textit{teacher}: Correct. 3 raised to the third power is 3 times 3 times 3, which is 27. Let's try another one.",

\textbf{Ask a question (Intention: Correct the student's mistake)}
\textit{Lesson}: Identify Unit Rate\\
\textit{student}: ok \\
\textit{tutor}: We can find the unit rate by selecting one point on the line and writing down the coordinates.\\
\textit{tutor}: What is the coordinate of one point on the line?\\
\textit{student}: 300\\
\textit{teacher}: Great try! Remember that for coordinates we need an x and a y value. Which one is 300? \\
\textit{student (simulated)}: x\\
\textit{teacher}: Great job - now what about y? \\

\paragraph{Provide a hint (Intention: Hint at the student's mistake)}

\textit{Lesson}: Multiplying Multi-Digit by One-Digit Numbers \\
\textit{tutor}: Give it your best shot. \\
\textit{tutor}: Good going!\\
\textit{tutor}: Is that your final answer?\\
\textit{student}: yes\\
\textit{teacher}: Great try! Remember when multiplying that we need to carry the tens place\\
\textit{student (simulated)}: o yah. So I carry the 2?\\
\textit{teacher}: Exactly!\\
    
\paragraph{Provide a strategy (Intention: Support the student in their thinking or problem-solving)}

\textit{Lesson}: Metric Conversions and Word Problems\\
\textit{student}: "yes i understand\\
\textit{tutor}: Let's move on to the next question. \\
\textit{tutor}: How many milliliters are in one liter? \\
\textit{student}: is that a 20 \\
\textit{teacher}:  Great try! Try using the prefixes to help you. Milli means thousand. \\
\textit{student (simulated)}:  oh then thousand?\\
\textit{teacher}: "Yes! There are 1000 milliliters in a liter.\\

\paragraph{Provide a worked example (Intention: Help the student understand the lesson topic or solution strategy)}

\textit{Lesson}: Understanding Multiplication\\
\textit{tutor}: Go ahead and give your best. \\
\textit{student}: c\\
\textit{tutor}:  Is that your final answer? \\
\textit{student}: yes \\
\textit{teacher}:  That's a great try! When we are multiplying, try to think of x groups of y and how many that would leave you. For example, 3 times 5 would be 3 groups of 5 or 5 groups of 3, which is 15. \\
\textit{teacher}: So, what would 2 times 4 be?
\textit{student (simulated)}: 8? \\
\textit{teacher}: Great job! \\

\paragraph{Provide a minor correction (Intention: Support the student in their thinking or problem-solving)}

\textit{Lesson}: Decomposing Figures\\
\textit{tutor}:  What is the area of the bottom rectangle?\\
\textit{tutor}: That's correct\\
\textit{tutor}: What is the area of the top rectangle?\\
\textit{student}: es 50\\
\textit{teacher}:  Ah, not quite. What is the length and the width of the top rectangle?\\
\textit{student (simulated)}: 2 and 5\\
\textit{teacher}: Yes. Do you remember how to get the area of a rectangle?\\
\textit{student (simulated)}: we multiply\\
\textit{teacher}: Yes, please multiply 2 x 5\\
\textit{student (simulated)}: 10\\

\paragraph{Provide a similar problem (Intention: Help the student understand the lesson topic or solution strategy)}

\textit{Lesson}: Dividing Using Equations\\
\textit{tutor}: Try to answer my question.\\
\textit{tutor}: Okay, let me explain it to you.\\
\textit{tutor}: What is the value of 5 times 2?\\
\textit{student}: 5\\
\textit{teacher}: Great try! 5 times 1 is 5. What about 5 times 2?\\
\textit{student (simulated)}: twice. so 10 \\
\textit{teacher}: Correct!\\

\paragraph{Simplify the question (Intention: Help the student understand the lesson topic or solution strategy)}

\paragraph{Affirm the correct answer (Intention: Signal to the student that they have solved or not solved the problem)}

\textit{Lesson}: Shapes and Area \\
\textit{tutor}:  I was facing some technical glitch.\\
\textit{student}: ya\\
\textit{tutor}:  What is area?\\
\textit{student}: spase in side sompthing\\
\textit{teacher}: Yes well done! Area is the 2D space inside something \\
\textit{student (simulated)}: ok\\
\textit{teacher}: Surface area is the area on the outside of a 3D object\\

\paragraph{Encourage the student (Intention: Motivate the student)}

\textit{Lesson}: Rounding\\
\textit{tutor}:  Here, the value of 6 in the given number is more than 5.\\
\textit{tutor}: So, we need to round the value 7.\\
\textit{tutor}: Do you have any questions about that part?\\
\textit{student}: millions\\
\textit{teacher}: Ok, ask away!\\
\textit{student (simulated)}: why do we round up?\\
\textit{teacher}: Becuase the 6 is greater than 5 (5 is the cutoff)\\

\section{Data Processing and Annotation \label{app:data_collection}}

This section discusses how the initial dataset is processed and how the dataset is annotated.

\subsection{Data Use}
The research team executed Data Use Agreements with both the tutoring provider and school district that outlined the allowable usage of the data to improve instruction in collaboration with an educational agency. 
Following the FERPA guidelines, we were eligible to engage in secondary data analysis with student data, which is what we did for this study. 
This study falls under the research team’s IRB for conducting research in collaboration with tutoring providers and school district (Protocol \#XXXX - redacted due to anonymous submission). 

\subsection{Data Processing}

\textbf{Signalling Expressions for Student Mistakes}
The following is the list of the signalling expressions used by the tutor which we use to mark conversation segments where the student has made a mistake. 
To identify these segments, we first lowercase all the conversation utterances, and check whether the following expressions exactly occur in the conversation. 

\begin{itemize}
    \item ``incorrect''
    \item ``not quite''
    \item ``bit off''
    \item ``good try''
    \item ``great try''
    \item ``effort''
    \item ``recheck''
\end{itemize}

\subsection{Annotation Quality Check}

We perform quality checks before the teachers started annotation. 
First, they are onboarded by an author of this work through two meetings, each meeting ranging between 30-60 minutes.
After the meeting, the teachers complete a sample of 20 problems similar to the ones in the final task. 
The teachers and author then meet again to walk through their answers and check their understanding of each of the taxonomy's category options.
The 20 sample problems are not used for the dataset and are only for onboarding purposes.
After training, each item took about 2 to 10 minutes for the teachers to complete.  

\subsection{Annotation Setup}

\begin{figure*}[t]
    \centering
    \begin{subfigure}[b]{0.49\textwidth}
        \centering
        \includegraphics[width=\textwidth]{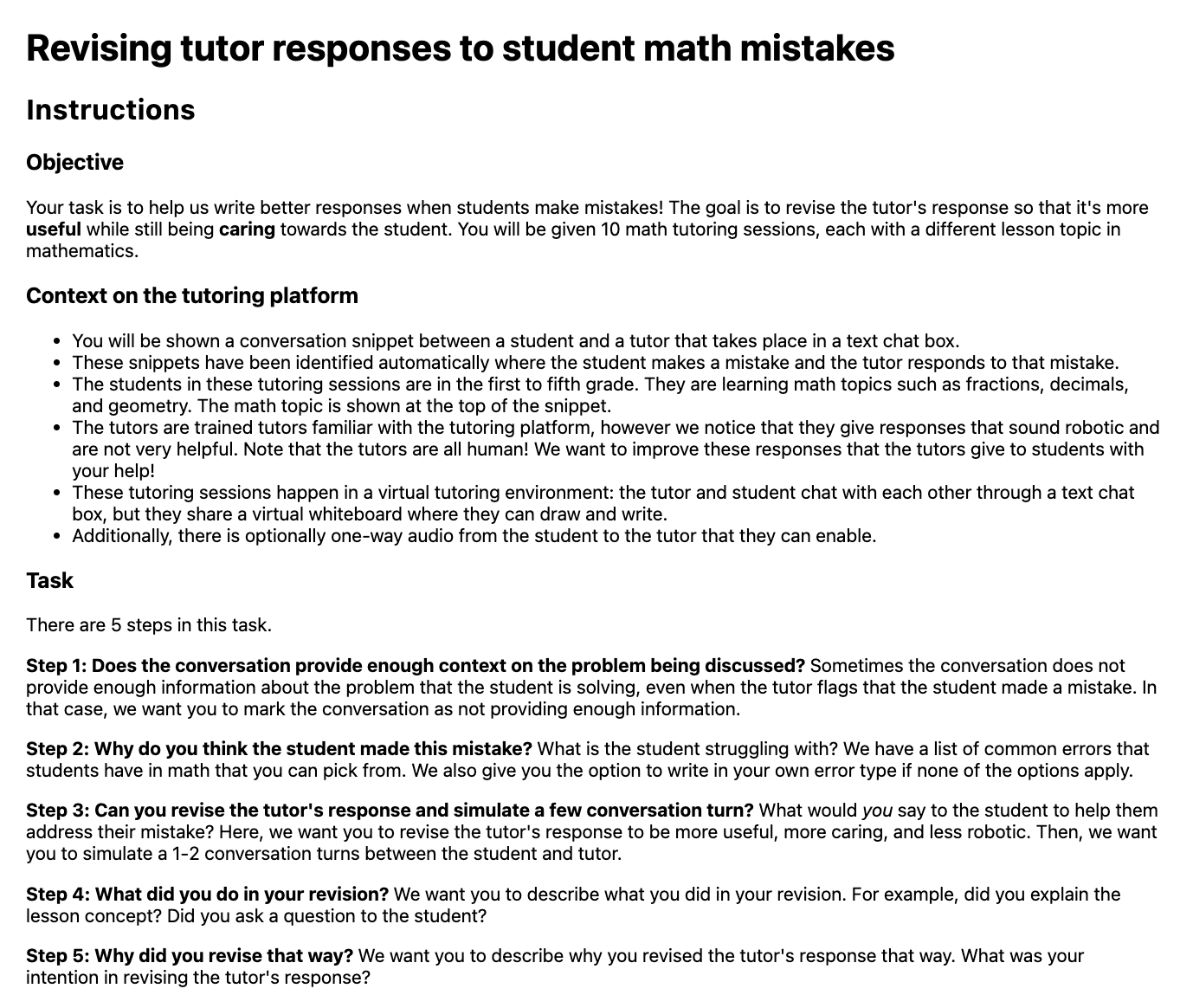}
        \caption{Instructions} 
    \end{subfigure}
    \hfill
    \begin{subfigure}[b]{0.49\textwidth}  
        \centering 
        \includegraphics[width=\textwidth]{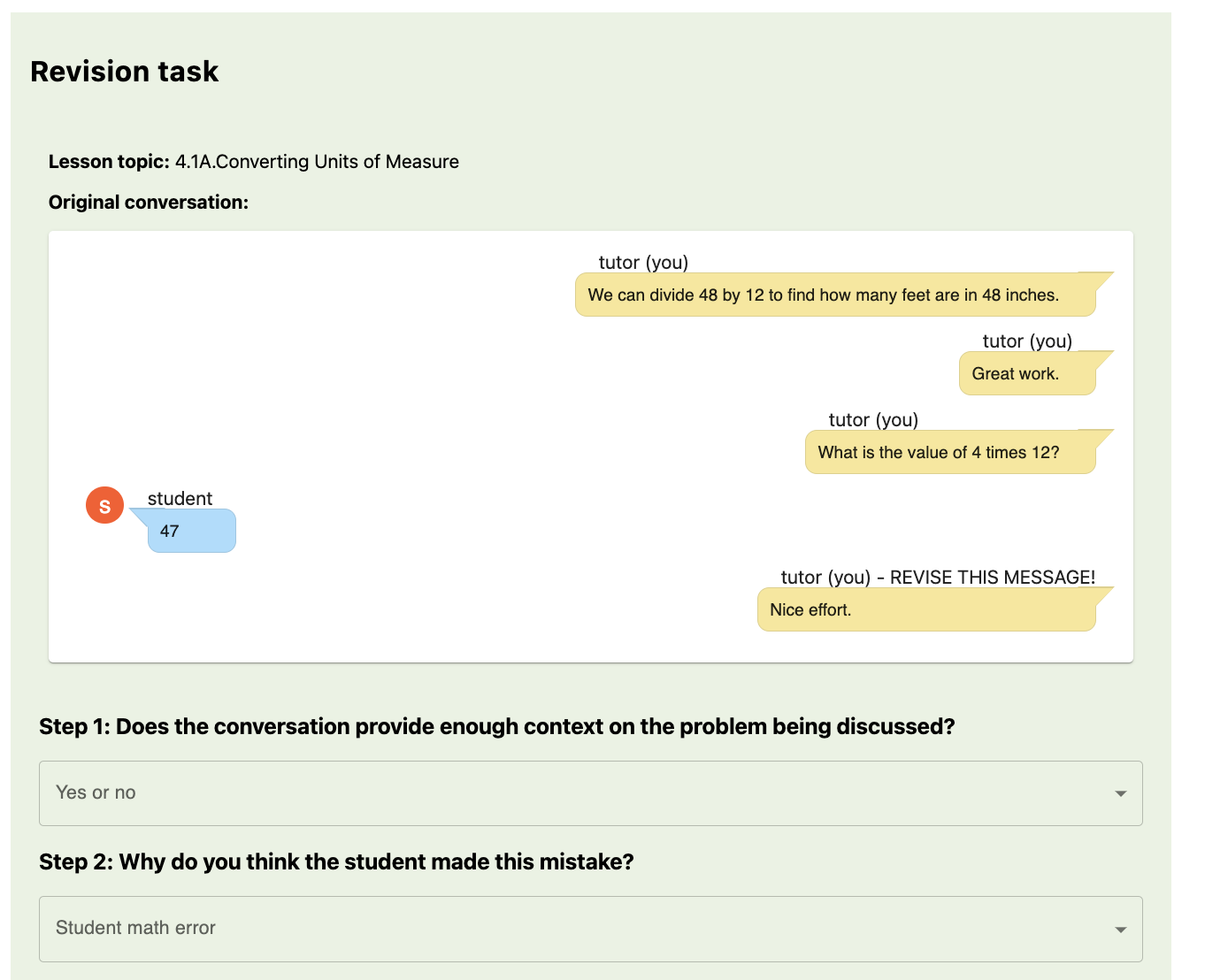}
        \caption{Step 1 \& 2} 
    \end{subfigure}

    \begin{subfigure}[b]{0.49\textwidth}  
        \centering 
        \includegraphics[width=\textwidth]{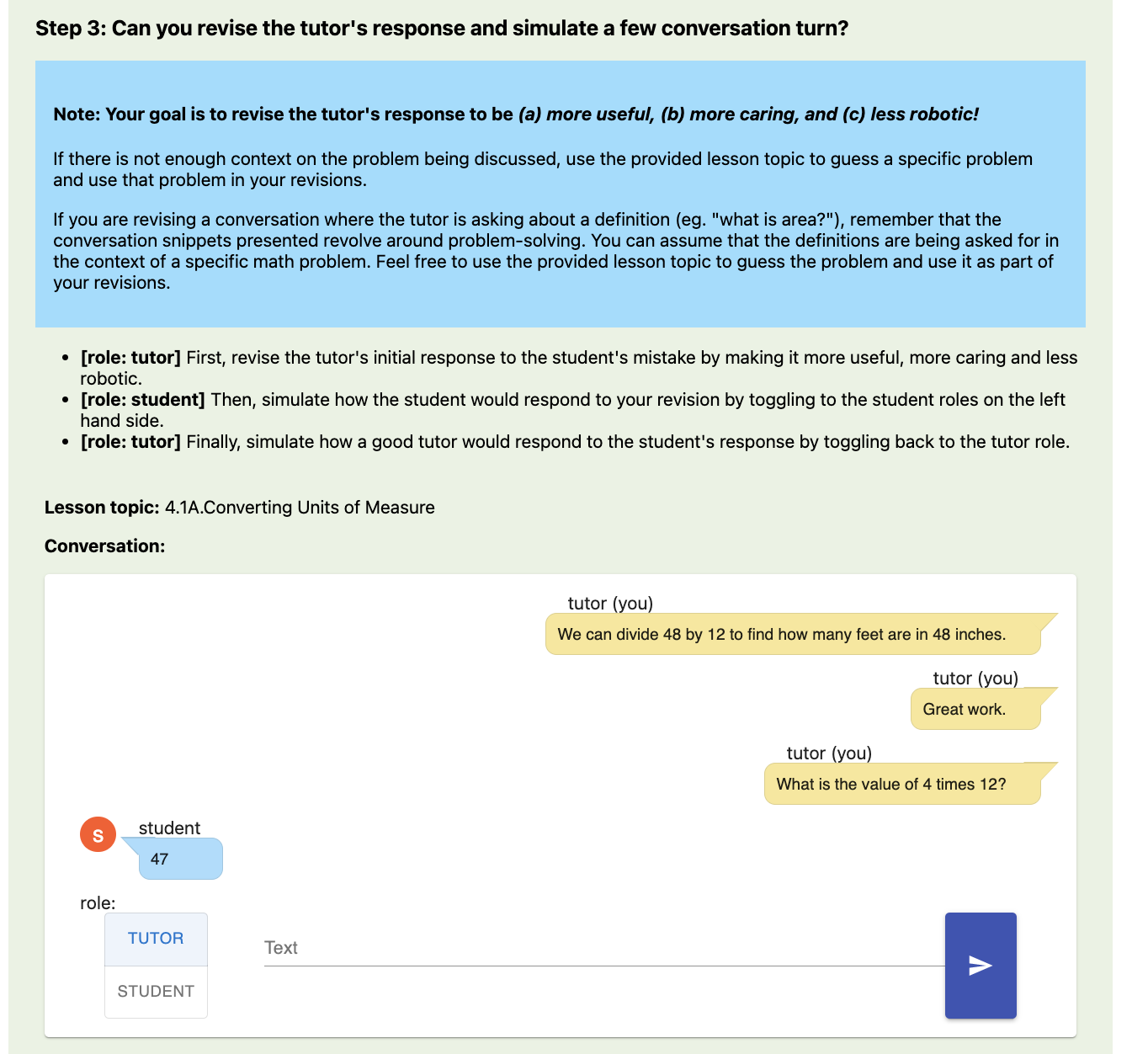}
        \caption{Step 3} 
    \end{subfigure}
    \begin{subfigure}[b]{0.49\textwidth}  
        \centering 
        \includegraphics[width=\textwidth]{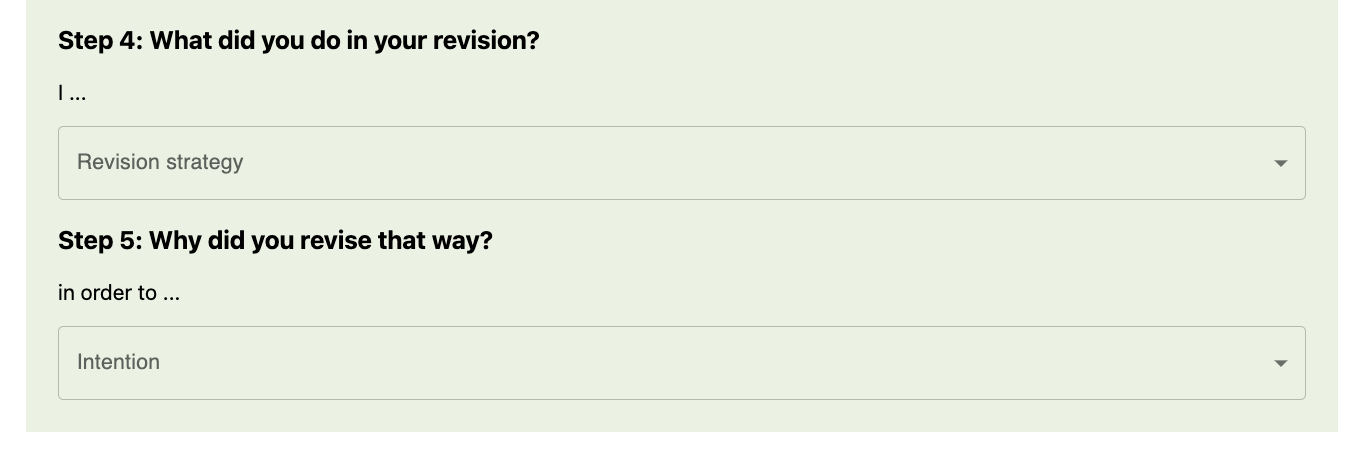}
        \caption{Step 4 \& 5} 
    \end{subfigure}
    \caption{Annotation interface for collecting decisions and responses.}
    \label{fig:revision_annotation_interface}
\end{figure*}

Figure~\ref{fig:revision_annotation_interface} shows the interface used by the teachers for annotating the data in our dataset.
Note that the annotation interface allows teachers to simulate the student's response. 
We have this feature because the teachers found that only responding on a single turn was not sufficient for them to complete their strategy of choice. 

\section{Prompts \label{app:prompting}}

This section contains information on the prompts for \gptfour{}, \chatgpt{}, and \llama{}.
We found that we could use similar prompts for \gptfour{} and \chatgpt{}, however these prompts had to adapted for \llama{} to mimic its training format\footnote{\url{https://gpus.llm-utils.org/llama-2-prompt-template/\#notes}, \url{https://huggingface.co/blog/llama2\#how-to-prompt-llama-2}}.
Unless otherwise noted, our prompt practices follow a mix of works from NLP, education and social sciences \citep{inverse2022, minichain2023, ziems2023can, wang2023sight}. 
For generating the remediation response, we found it important to add a length constraint to force the model to stick to the short message styles of the tutor and student; otherwise, the model responses would generally be extremely long (up to $5-10\times$ longer than the original tutor responses).
Adding the length constraint also prevented the model from simulating the rest of the tutoring session. 
All the prompts include context on the task at the start of the prompt, and the constraints of outputting a JSON-formatted text for the task at the end of the prompt.

\subsection{No Decision-Making Condition}
Models directly respond, $c_r \sim p(c_r|c_h)$. 
The prompts for \gptfour{} and \chatgpt{} are shown in Figure~\ref{fig:no_decision_making_prompt_openai}.
The prompt for \llama{} is shown in Figure~\ref{fig:no_decision_making_prompt_llama} where the formatting is slightly adapted.

\begin{figure*}[h]
    \centering
    \small
    \begin{tcolorbox}[
    taskc,
    title={\textbf{No Decision-Making Prompt for \gptfour{} and \chatgpt{}}},
    ]
    You are an experienced elementary math teacher and you are going to respond to a student's mistake in a useful and caring way. The problem your student is solving is on topic: \{lesson\_topic\}.\\
\{c\_h\}\\
tutor (maximum one sentence):
    \end{tcolorbox}
    \caption{
    \textbf{Prompt for the \textit{no decision-making} condition for \gptfour{} and \chatgpt{}.}
    \texttt{\{lesson\_topic\}} is the placeholder for the lesson topic discussed in the conversation. 
    \texttt{\{c\_h\}} is the placeholder for the conversation history leading up to (and including) the student's message that contains the mistake.  
    We add an additional constraint ``(maximum one sentence)'' because from our experiments, \chatgpt{} and \gptfour{} typically output extremely long responses that would be unnatural for this tutoring conversation domain.
    \label{fig:no_decision_making_prompt_openai}}
\end{figure*}

\begin{figure*}[h]
    \centering
    \small
    \begin{tcolorbox}[
    taskc,
    title={\textbf{No Decision-Making Prompt for \llama{}}},
    ]
    \#\#\# System:\\
You are an experienced elementary math teacher and you are going to respond to a student's mistake in a useful and caring way.\\ \\
\#\#\# User: \\
Lesson topic: \{lesson\_topic\} \\
Conversation:\\
\{c\_h\} \\ \\
\#\#\# Assistant: \\
tutor (maximum one sentence):
    \end{tcolorbox}
    \caption{
    \textbf{Prompt for the \textit{no decision-making} condition for \llama{}.}
    \texttt{\{lesson\_topic\}} is the placeholder for the lesson topic discussed in the conversation. 
    \texttt{\{c\_h\}} is the placeholder for the conversation history leading up to (and including) the student's message that contains the mistake.  
    \label{fig:no_decision_making_prompt_llama}}
\end{figure*}

\subsection{Expert Decision-Making Condition \label{app:expert_decision_making}}
 Models generate with the expert's decisions,  $c_r \sim p(c_r|c_h, e, z_{\text{what}}, z_{\text{why}})$.
The prompts for \gptfour{} and \chatgpt{} are shown in Figure~\ref{fig:expert_decision_making_prompt_openai}.
The prompt for \llama{} is shown in Figure~\ref{fig:expert_decision_making_prompt_llama} where the formatting is slightly adapted.
The labels for $e, z_{\text{what}}, z_{\text{why}}$ come from our annotated dataset.

\begin{figure*}[h]
    \centering 
    \small
    \begin{tcolorbox}[
    taskc,
    title={\textbf{Decision-Making Prompt for \gptfour{} and \chatgpt{}}},
    ]
    You are an experienced elementary math teacher and you are going to respond to a student's mistake in a useful and caring way. The problem your student is solving is on topic: \{lesson\_topic\}. \{e\} \{z\_what\} in order to \{z\_why\}. \\
\{c\_h\}\\
tutor (maximum one sentence):
    \end{tcolorbox}
    \caption{
    \textbf{Prompt for the \textit{decision-making} condition for \gptfour{} and \chatgpt{}.}
    \texttt{\{lesson\_topic\}} is the placeholder for the lesson topic discussed in the conversation. 
    The error, strategy, and intention decisions are included in the prompt where \texttt{\{e\}} is a placeholder for the error type, \texttt{\{z\_what\}} for the strategy and \texttt{\{z\_why\}} for the intention. 
    Note that each of the decisions are formatted to be a coherent piece of text. 
    \texttt{\{c\_h\}} is the placeholder for the conversation history leading up to (and including) the student's message that contains the mistake.  
    We add an additional constraint ``(maximum one sentence)'' because from our experiments, \chatgpt{} and \gptfour{} typically output extremely long responses that would be unnatural for this tutoring conversation domain.
    \label{fig:expert_decision_making_prompt_openai}}
\end{figure*}

\begin{figure*}[h]
    \centering
    \small
    \begin{tcolorbox}[
    taskc,
    title={\textbf{Decision-Making Prompt for \llama{}}},
    ]
    \#\#\# System:\\
You are an experienced elementary math teacher and you are going to respond to a student's mistake in a useful and caring way.\\ \\
\#\#\# User: \\
\{e\} \{z\_what\} in order to \{z\_why\}. \\ 
Lesson topic: \{lesson\_topic\} \\
Conversation:\\
\{c\_h\} \\ \\
\#\#\# Assistant: \\
tutor (maximum one sentence):
    \end{tcolorbox}
    \caption{
    \textbf{Prompt for the \textit{decision-making} condition for \llama{}.}
    \texttt{\{lesson\_topic\}} is the placeholder for the lesson topic discussed in the conversation. 
    The error, strategy, and intention decisions are included in the prompt where \texttt{\{e\}} is a placeholder for the error type, \texttt{\{z\_what\}} for the strategy and \texttt{\{z\_why\}} for the intention. 
    Note that each of the decisions are formatted to be a coherent piece of text. 
    \texttt{\{c\_h\}} is the placeholder for the conversation history leading up to (and including) the student's message that contains the mistake.  
    \label{fig:expert_decision_making_prompt_llama}}
\end{figure*}

\subsection{Self decision-making condition}
LLMs make their own decisions, then generate responses based on them, $c_r \sim p(c_r|c_h, e^{\text{model}}, z^{\text{model}}_{\text{what}}, z^{\text{model}}_{\text{why}})$. 
Following the decision-making model, we first generate the model's decision on error $e^{\text{model}}$ with prompts in Figure~\ref{fig:error_decision_making_prompt_openai} (for \gptfour{} and \chatgpt{}) and in Figure~\ref{fig:error_decision_making_prompt_llama} (for \llama{}). 
Then we generate the model's decision on strategy and intention $z^{\text{model}}_{\text{what}}, z^{\text{model}}_{\text{why}}$ in Figure~\ref{fig:z_decision_making_prompt_openai} (for \gptfour{} and \chatgpt{}) and in Figure~\ref{fig:z_decision_making_prompt_llama} (for \llama{}). 
Finally, we use the previous response generation prompts with decision-making to generate $c_r$ from Section~\ref{app:expert_decision_making}.

\begin{figure*}[h]
    \centering 
    \small
    \begin{tcolorbox}[
    taska,
    title={\textbf{Determine Error ($e$) with \gptfour{} and \chatgpt{}.}},
    ]
    You are an experienced elementary math teacher. Your task is to read a conversation snippet of a tutoring session between a student and tutor, and determine what type of error the student makes in the conversation. We have a list of common errors that students make in math, which you can pick from. We also give you the option to write in your own error type if none of the options apply. \\

Error list: \\ 
0. Student does not seem to understand or guessed the answer. \\
1. Student misinterpreted the question. \\
2. Student made a careless mistake. \\
3. Student has the right idea, but is not quite there. \\
4. Student's answer is not precise enough or the tutor is being too picky about the form of the student's answer. \\
5. None of the above, but I have a different description (please specify in your reasoning). \\
6. Not sure, but I'm going to try to diagnose the student. \\

Here is the conversation snippet: \\
Lesson topic: \{lesson\_topic\} \\
Conversation: \\
\{c\_h\} \\

Why do you think the student made this mistake? Pick an option number from the error list and provide the reason behind your choice. Format your answer as: [\{"answer": \#, "reason": "write out your reason for picking \# here"\}]
    \end{tcolorbox}
    \caption{
    \textbf{Prompt to determine error $e$ with \gptfour{} and \chatgpt{}.} \texttt{\{lesson\_topic\}} is the placeholder for the lesson topic discussed in the conversation. \texttt{\{c\_h\}} is the placeholder for the conversation history leading up to (and including) the student's message that contains the mistake.  
    \label{fig:error_decision_making_prompt_openai}}
\end{figure*}

\begin{figure*}[h]
    \centering 
    \small
    \begin{tcolorbox}[
    taska,
    title={\textbf{Determine Error ($e$) with \llama{}.}},
    ]
    \#\#\# System: \\
    You are an experienced elementary math teacher. Your task is to read a conversation snippet of a tutoring session between a student and tutor, and determine what type of error the student makes in the conversation. We have a list of common errors that students make in math, which you can pick from. We also give you the option to write in your own error type if none of the options apply. \\

Error list: \\ 
0. Student does not seem to understand or guessed the answer. \\
1. Student misinterpreted the question. \\
2. Student made a careless mistake. \\
3. Student has the right idea, but is not quite there. \\
4. Student's answer is not precise enough or the tutor is being too picky about the form of the student's answer. \\
5. None of the above, but I have a different description (please specify in your reasoning). \\
6. Not sure, but I'm going to try to diagnose the student. \\

Format your answer as: [\{"answer": \#, "reason": "write out your reason for picking \# here"\}]\\ \\
\#\#\# User: \\
Lesson topic: \{lesson\_topic\} \\
Conversation: \\
\{c\_h\} \\
\\
\#\#\# Assistant: \\
\text{[\{}"answer": 
    \end{tcolorbox}
    \caption{
    \textbf{Prompt to determine error $e$ with \llama{}.} 
    \texttt{\{lesson\_topic\}} is the placeholder for the lesson topic discussed in the conversation. \texttt{\{c\_h\}} is the placeholder for the conversation history leading up to (and including) the student's message that contains the mistake.  
    \label{fig:error_decision_making_prompt_llama}}
\end{figure*}

\begin{figure*}[h]
    \centering 
    \small
    \begin{tcolorbox}[
    taskb,
    title={\textbf{Determine Strategy and Intention ($z_{\text{what}}, z_{\text{why}}$) with \gptfour{} and \chatgpt{}.}},
    ]
    You are an experienced elementary math teacher. Your task is to read a conversation snippet of a tutoring session between a student and tutor where a student makes a mistake. You should then determine what strategy you want to use to remediate the student's error, and state your intention in using that strategy. We have a list of common strategies and intentions that teachers use, which you can pick from. We also give you the option to write in your own strategy or intention if none of the options apply. \\

Strategies: \\
0. Explain a concept \\
1. Ask a question \\
2. Provide a hint \\
3. Provide a strategy \\
4. Provide a worked example \\
5. Provide a minor correction \\
6. Provide a similar problem \\
7. Simplify the question \\
8. Affirm the correct answer \\
9. Encourage the student \\
10. Other (please specify in your reasoning) \\

Intentions: \\
0. Motivate the student \\
1. Get the student to elaborate their answer \\
2. Correct the student's mistake \\
3. Hint at the student's mistake \\
4. Clarify a student's misunderstanding \\
5. Help the student understand the lesson topic or solution strategy \\
6. Diagnose the student's mistake \\
7. Support the student in their thinking or problem-solving \\
8. Explain the student's mistake (eg. what is wrong in their answer or why is it incorrect) \\
9. Signal to the student that they have solved or not solved the problem \\
10. Other (please specify in your reasoning) \\

Here is the conversation snippet: \\
Lesson topic: \{lesson\_topic\} \\
Conversation: \\
\{c\_h\} \\

How would you remediate the student's error and why? Pick the option number from the list of strategies and intentions and provide the reason behind your choices. Format your answer as: [\{"strategy": \#, "intention": \#, "reason": "write out your reason for picking that strategy and intention"\}]
    \end{tcolorbox}
    \caption{
    \textbf{Prompt to determine strategy and intention $z_{\text{what}}, z_{\text{why}}$ with \gptfour{} and \chatgpt{}.} \texttt{\{lesson\_topic\}} is the placeholder for the lesson topic discussed in the conversation. \texttt{\{c\_h\}} is the placeholder for the conversation history leading up to (and including) the student's message that contains the mistake.  
    \label{fig:z_decision_making_prompt_openai}}
\end{figure*}

\begin{figure*}[h]
    \centering 
    \small
    \begin{tcolorbox}[
    taskb,
    title={\textbf{Determine Strategy and Intention ($z_{\text{what}}, z_{\text{why}}$) with \llama{}.}},
    ]
    \#\#\# System: \\
    You are an experienced elementary math teacher. Your task is to read a conversation snippet of a tutoring session between a student and tutor where a student makes a mistake. You should then determine what strategy you want to use to remediate the student's error, and state your intention in using that strategy. We have a list of common strategies and intentions that teachers use, which you can pick from. We also give you the option to write in your own strategy or intention if none of the options apply.\\

Strategies: \\
0. Explain a concept \\
1. Ask a question \\
2. Provide a hint \\
3. Provide a strategy \\
4. Provide a worked example \\
5. Provide a minor correction \\
6. Provide a similar problem \\
7. Simplify the question \\
8. Affirm the correct answer \\
9. Encourage the student \\
10. Other (please specify in your reasoning) \\

Intentions: \\
0. Motivate the student \\
1. Get the student to elaborate their answer \\
2. Correct the student's mistake \\
3. Hint at the student's mistake \\
4. Clarify a student's misunderstanding \\
5. Help the student understand the lesson topic or solution strategy \\
6. Diagnose the student's mistake \\
7. Support the student in their thinking or problem-solving \\
8. Explain the student's mistake (eg. what is wrong in their answer or why is it incorrect) \\
9. Signal to the student that they have solved or not solved the problem \\
10. Other (please specify in your reasoning) \\

Format your answer as: [\{"strategy": \#, "intention": \#, "reason": "write out your reason for picking \# here"\}]\\ \\
\#\#\# User: \\
Lesson topic: \{lesson\_topic\} \\
Conversation: \\
\{c\_h\} \\
\\
\#\#\# Assistant: \\
\text{[\{}"strategy": 
    \end{tcolorbox}
    \caption{
    \textbf{Prompt to determine error $z_{\text{what}}, z_{\text{why}}$ with \llama{}.} 
    \texttt{\{lesson\_topic\}} is the placeholder for the lesson topic discussed in the conversation. \texttt{\{c\_h\}} is the placeholder for the conversation history leading up to (and including) the student's message that contains the mistake.  
    \label{fig:z_decision_making_prompt_llama}}
\end{figure*}

\subsection{Random Decision-Making Condition}
We randomly select a decision for the error, strategy and intention. Then, we use the previous response generation prompts with decision-making to generate $c_r$ from Section~\ref{app:expert_decision_making}.

\section{Human Evaluations \label{app:human_evaluation_taskc} }
We describe the human evaluation setup, whose results are reported in Section~\ref{sec:main_results_task_c}.

The human evaluations were run on Prolific.
Our prescreening criteria were that the participants have to be located in the USA,  have to be a teacher, their fluent languages have to include English, and their approval rating has to be at least 96\%.
We conduct the human evaluations on 40 items from each model with 3 raters; 10 of these items were held to be the same and the other 30 were randomly sampled. 
The 10 items are used to calculated the IRR reported in the main tables. 
Each item consisted of a pair of remediation responses, Response A and Response B. 
One of the responses is the original tutor's response to the student's mistake, and the other response is the newly generated remediation response (ie. the expert-written response in the \texttt{Human} row, and the model-generated response in the other rows).
The ordering of the responses is always randomized. 
Each item is scored on a Likert scale from -2 to 2 on four dimensions: \textit{usefulness}, \textit{care}, \textit{human-soundingness}, and \textit{preference}. 
We also provided a definition for each dimension.

\begin{figure*}[t]
    \centering
    \begin{subfigure}[b]{0.49\textwidth}
        \centering
        \includegraphics[width=\textwidth]{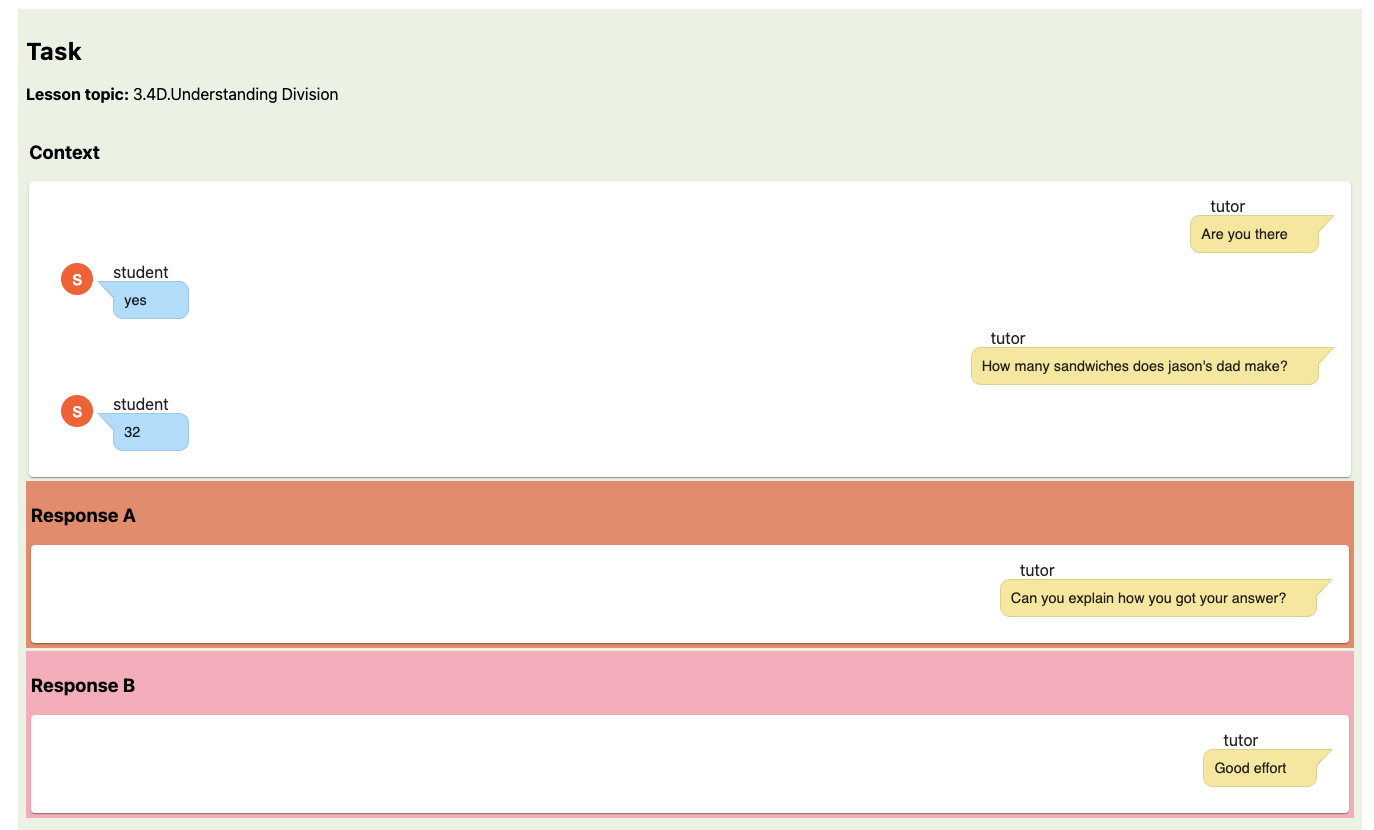}
    \end{subfigure}
    \hfill
    \begin{subfigure}[b]{0.49\textwidth}  
        \centering 
        \includegraphics[width=\textwidth]{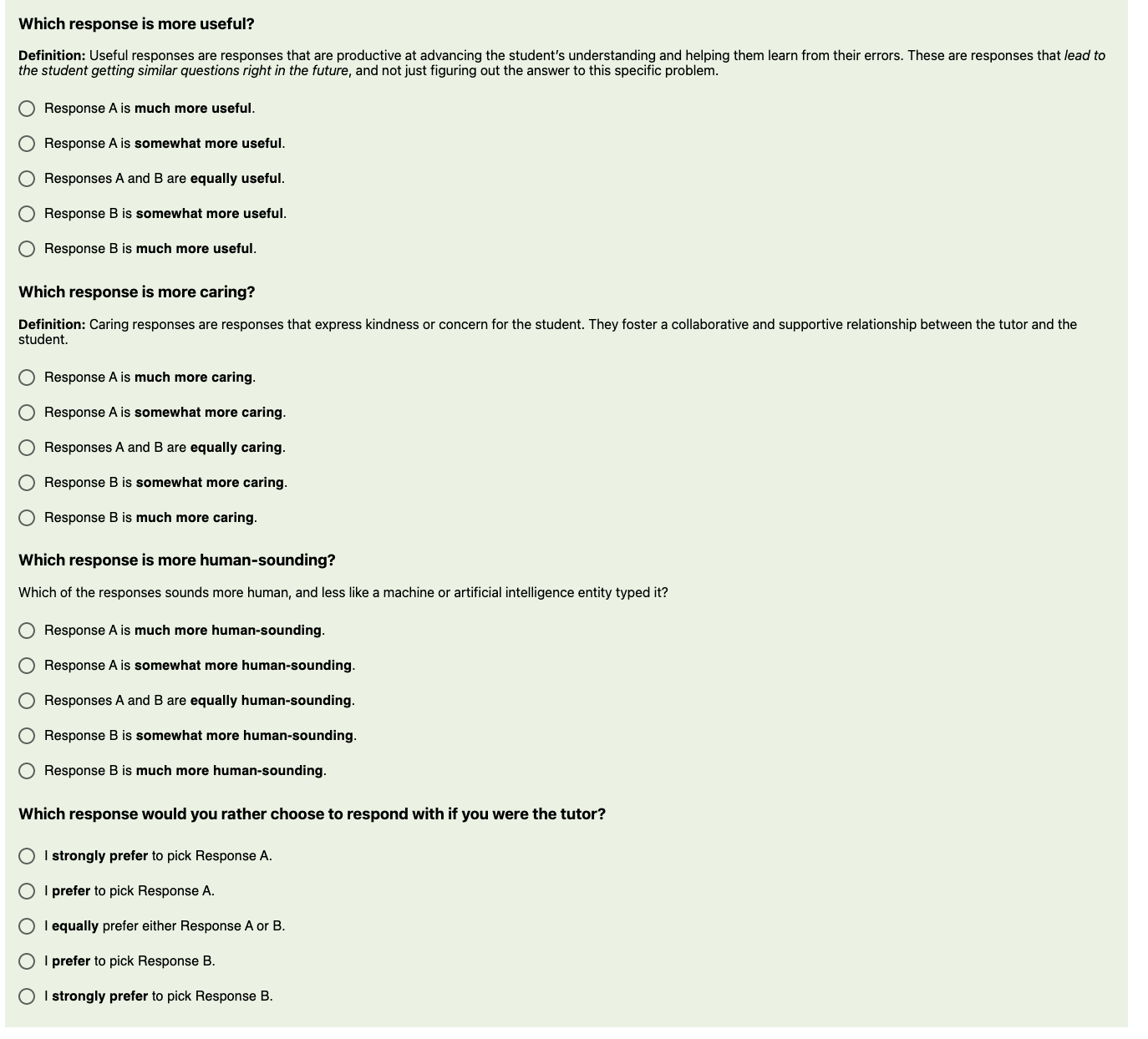}
    \end{subfigure}
    \caption{Annotation interface for evaluating the remediation responses.}
    \label{fig:evaluation_task_c_annotation_interface}
\end{figure*}

Figure~\ref{fig:evaluation_task_c_annotation_interface} shows an example of the evaluation interface. 
Specifically, the phrasing for each dimension was:

\textbf{Which response is more useful?} \\
\textit{Definition:} Useful responses are responses that are productive at advancing the student’s understanding and helping them learn from their errors. These are responses that lead to the student getting similar questions right in the future, and not just figuring out the answer to this specific problem.

\begin{itemize}
    \item Response A is much more useful.
    \item Response A is somewhat more useful.
    \item  Responses A and B are equally useful.
    \item Response B is somewhat more useful.
    \item Response B is much more useful.
\end{itemize}

\textbf{Which response is more caring?} \\
\textit{Definition:} Caring responses are responses that express kindness or concern for the student. They foster a collaborative and supportive relationship between the tutor and the student.

\begin{itemize}
    \item Response A is much more caring.
    \item  Response A is somewhat more caring.
    \item Responses A and B are equally caring.
    \item Response B is somewhat more caring.
    \item Response B is much more caring.
\end{itemize}

\textbf{Which response is more human-sounding?} \\
Which of the responses sounds more human, and less like a machine or artificial intelligence entity typed it?

\begin{itemize}
    \item Response A is much more human-sounding.
    \item  Response A is somewhat more human-sounding.
    \item  Responses A and B are equally human-sounding.
    \item  Response B is somewhat more human-sounding.
    \item Response B is much more human-sounding.
\end{itemize}

\textbf{Which response would you rather choose to respond with if you were the tutor?}

\begin{itemize}
    \item I strongly prefer to pick Response A.
    \item I prefer to pick Response A.
    \item I equally prefer either Response A or B.
    \item I prefer to pick Response B.
    \item I strongly prefer to pick Response B.
\end{itemize}

\section{Lexical analysis \label{app:lexical_analysis}}

Table~\ref{tab:lexical_chatgpt} compares the top-5 bigram usage for \texttt{ChatGPT} in all decision-making conditions. 
Table~\ref{tab:lexical_llama} does the same for \texttt{Llama-2-70b-instruct}.

\begin{table*}[t]
      \centering
      \resizebox{\textwidth}{!}{%
        \def\arraystretch{1.15}
      \begin{tabular}{cc|cc|cc|cc}
        \toprule
       \multicolumn{2}{c}{\bf \texttt{None} + \texttt{ChatGPT}} & \multicolumn{2}{c}{\bf \texttt{Expert} + \texttt{ChatGPT}}  & \multicolumn{2}{c}{\bf \texttt{ChatGPT} + \texttt{ChatGPT}} & \multicolumn{2}{c}{\bf \texttt{Random} +\texttt{ChatGPT}} \\
       \midrule
       bigram & log odds & bigram & log odds & bigram & log odds & bigram & log odds \\
       \midrule
       effort\_remember & 2.34 & \hlc[cyan!30]{can\_explain}& 2.14 & \hlc[orange!50]{actually\_correct}   & 2.9 & thats\_close & 1.85 \\
       lets\_focus   & 1.32 & great\_start & 1.94 & \hlc[orange!50]{correct\_answer}     & 1.96 & example\_help & 1.69 \\
       carry\_tens    & 1.31 & can\_tell   & 1.88 & job\_attempting   & 1.42 & can\_think & 1.51 \\
       focus\_question & 1.31 & \hlc[cyan!30]{explain\_got} & 1.53 & small\_mistake   & 1.42 & can\_try & 1.51 \\
       clarify\_mean    & 1.31 & got\_answer & 1.42 & attempting\_problem     & 1.42 & glasses\_water & 1.47 \\
        \bottomrule
      \end{tabular}
      }
    \caption{Top 5 bigrams for \texttt{ChatGPT}. \textbf{\texttt{ChatGPT} with \textit{expert decision-making} engages more with the \hlc[cyan!30]{student's problem-solving process}, whereas \texttt{ChatGPT} with \textit{self decision-making} engages more with the \hlc[orange!50]{student's answer}}.
    \label{tab:lexical_chatgpt}}
\end{table*}





\begin{table*}[t]
      \centering
      \resizebox{\textwidth}{!}{%
        \def\arraystretch{1.15}
      \begin{tabular}{cc|cc|cc|cc}
        \toprule
       \multicolumn{2}{c}{\bf \texttt{None} + \texttt{Llama}} & \multicolumn{2}{c}{\bf \texttt{Expert} + \texttt{Llama}}  & \multicolumn{2}{c}{\bf \texttt{Llama} + \texttt{Llama}} & \multicolumn{2}{c}{\bf \texttt{Random} +\texttt{Llama}} \\
       \midrule
       bigram & log odds & bigram & log odds & bigram & log odds & bigram & log odds \\
       \midrule
       user\_lesson   & 8.02 & lets\_closer & 3.73  & user\_student & 5.79 & lets\_closer & 3.27 \\
       user\_tutor   & 4.27  & closer\_look & 3.73 & student\_responds & 4.65 & closer\_look & 3.27 \\
       respond\_students & 3.62  & look\_problem & 2.51 & student\_understand & 4.0 & right\_track & 2.49 \\
       mistake\_useful & 3.54 & look\_expression & 1.56 & response\_provide &  3.85 & youre\_right & 2.09 \\
       going\_respond  & 3.52 & groups\_objects & 1.56 & provide\_hint & 3.09 & look\_answer & 1.7 \\
        \bottomrule
      \end{tabular}
      }
    \caption{
    Top 5 bigrams for \texttt{Llama-2-70b-instruct}.
    \label{tab:lexical_llama}}
\end{table*}





\end{document}